\documentclass[11pt]{article}
\usepackage{booktabs}
\usepackage{acl}
\usepackage{subcaption}
\usepackage{makecell}
\usepackage{algorithm}
\usepackage{algpseudocode}
\floatname{algorithm}{Algorithm}

\usepackage[table]{xcolor}
\usepackage{booktabs, multirow}
\definecolor{highlightgray}{gray}{0.93} 
\usepackage{amssymb}   
\usepackage{amsfonts}  
\usepackage{times}
\usepackage{latexsym}
\usepackage{amsmath} 
\usepackage{tabularx}
\usepackage[T1]{fontenc}

\usepackage[utf8]{inputenc}

\definecolor{blue}{HTML}{00FFFF}

\usepackage{microtype}

\usepackage{inconsolata}

\usepackage{graphicx}

\title{Silence the Judge: Reinforcement Learning with Self-Verifier \\via Latent Geometric Clustering}

\author{
  \textbf{Nonghai Zhang}\textsuperscript{1,2}\thanks{Work done during internship at Meituan.} \thanks{Equal contribution.}
  \quad
  \textbf{Weitao Ma}\textsuperscript{1,3}\footnotemark[1]\footnotemark[2]
  \quad
  \textbf{Zhanyu Ma}\textsuperscript{1}
  \quad
  \textbf{Jun Xu}\textsuperscript{1}\thanks{Corresponding authors.}
  \quad\\
  \textbf{Jiuchong Gao}\textsuperscript{1}\footnotemark[3] 
  \quad
  \textbf{Jinghua Hao}\textsuperscript{1}
  \quad
  \textbf{Renqing He}\textsuperscript{1}
  \quad
  \textbf{Jingwen Xu}\textsuperscript{1}
\\
  \textsuperscript{1}Meituan, 
  \textsuperscript{2}Peking University, 
  \textsuperscript{3}Harbin Institute of Technology
\\
}

\begin{document}
\maketitle
\begin{abstract}

Group Relative Policy Optimization (GRPO) significantly enhances the reasoning performance of Large Language Models (LLMs). However, this success heavily relies on expensive external verifiers or human rules. Such dependency not only leads to significant computational costs and training latency, but also yields sparse rewards that hinder optimization efficiency. To address these challenges, we propose \textbf{Latent-GRPO}, a framework that derives intrinsic rewards directly from latent space geometry. Crucially, our empirical analysis reveals a compelling geometric property: terminal token representations of correct reasoning trajectories form dense clusters with high intra-class similarity, whereas incorrect trajectories remain scattered as outliers. In light of this discovery, we introduce the \textbf{Iterative Robust Centroid Estimation (IRCE)} algorithm, which generates dense, continuous rewards by mitigating magnitude fluctuations via spherical projection and estimating a robust ``truth centroid'' through iterative aggregation. Experimental results on multiple datasets show that our method maintains model performance while achieving a training speedup of over $2\times$ compared to baselines. Furthermore, extensive results demonstrate strong generalization ability and robustness. The code will be released soon.
\end{abstract}

\section{Introduction}

Large Language Models (LLMs) \cite{zhao2023survey} have achieved remarkable success in tackling complex reasoning tasks\cite{hendrycks2021measuring,cobbe2021training,chen2021evaluating}. To further enhance these capabilities, Reinforcement Learning from Human Feedback (RLHF) \cite{ouyang2022training} has been established as the standard paradigm for model alignment. Specifically, this process conventionally utilized Proximal Policy Optimization (PPO) \cite{rafailov2023direct} to refine policy performance. Furthermore, Group Relative Policy Optimization (GRPO) \cite{shao2024deepseekmath} simplifies the process by replacing the value model with group-based advantages to reduce computational costs.

\begin{figure}[t]
    \centering
    \includegraphics[width=\columnwidth]{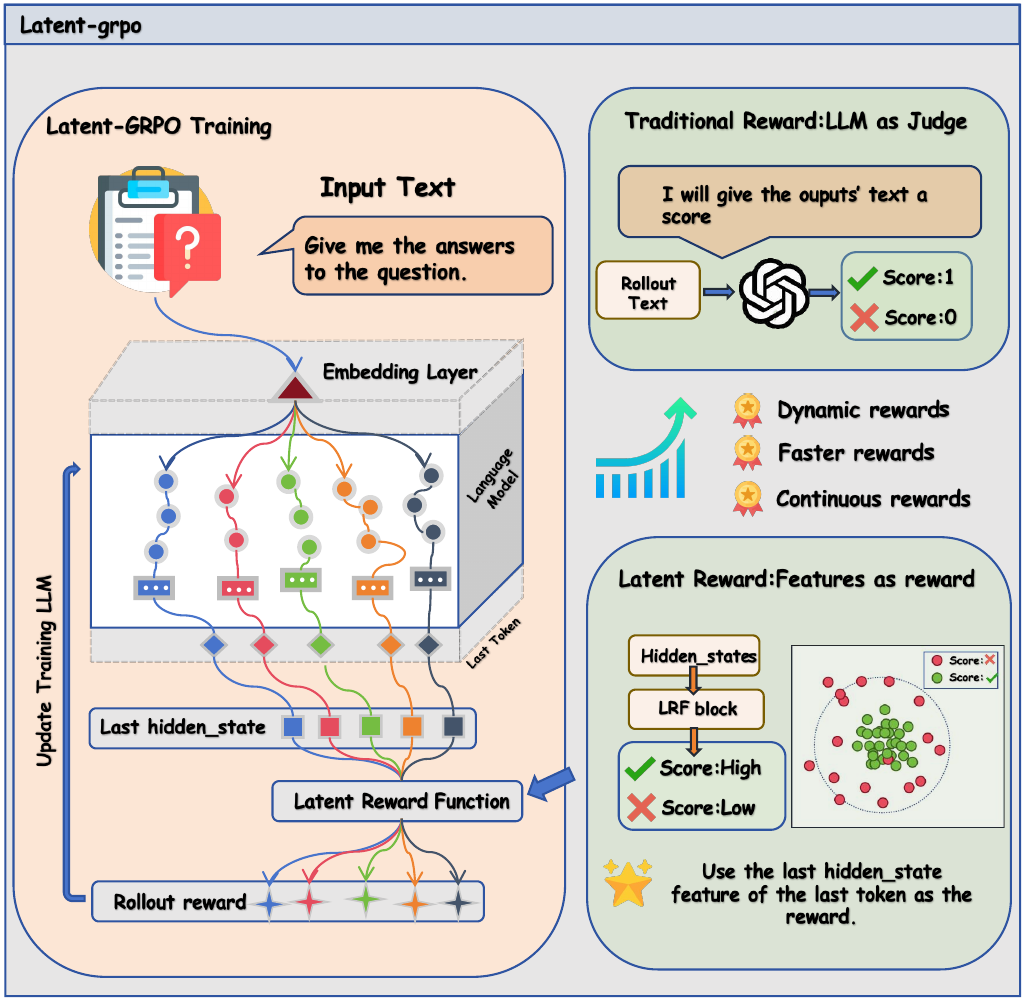}
    \caption{Comparison between conventional GRPO and Latent-GRPO. Conventional GRPO relies on expensive external verifiers to compute rewards, whereas Latent-GRPO autonomously extracts reward signals from the geometric structure of the latent space, eliminating external dependencies.}
    \label{fig:intro}
    \vspace{-0.4cm}

\end{figure}

However, the practical efficacy of GRPO is often constrained by its heavy reliance on external verifiers \cite{wen2025reinforcement, zheng2023judging, zhou2025reinforcing}, which makes training outcomes highly sensitive to their quality. On the one hand, rule-based verifiers are typically confined to deterministic tasks like mathematics. Moreover, designing clear and error-free rules for complex reasoning remains extremely difficult, and imperfect rules can severely degrade training performance. On the other hand, employing external LLMs or training additional reward models incurs substantial costs \cite{lightman2023let}. These approaches introduce significant computational overhead and inference latency, which ultimately slows down the entire training process. Furthermore, such external judges are susceptible to biases or inaccurate scoring, thereby compromising training stability and final model quality \cite{xu2025tinyv, cai2025reinforcement}. Beyond these concerns, most existing reward signals remain sparse and discrete \cite{tao2025hybrid}. This binary feedback fails to capture the continuous semantic nuances of the reasoning process, often leading the model toward reward hacking \cite{cui2025process, gao2023scaling}. To address these challenges, we argue that an ideal reward mechanism should be \textbf{intrinsic, dense, and training-free}.

Building on representation engineering \cite{zou2023representation, bartoszcze2025representation}, which reveals that LLMs encode high-level semantic concepts internally \cite{marks2023geometry}, we discover a striking geometric property: the \textbf{last hidden states of terminal tokens} in correct reasoning trajectories form dense clusters, while incorrect paths remain scattered. Theoretically, this stems from the Transformer's attention mechanism, which progressively aggregates the reasoning context into the final representation. Furthermore, this geometric consistency reflects the model's inherent discriminative capabilities acquired during large-scale pre-training \cite{radford2018improving}. In essence, the latent space acts as an implicit verifier where logical consistency manifests as semantic convergence, providing a robust foundation for intrinsic reward modeling.

We introduce \textbf{Latent-GRPO}, a framework that utilizes robust intrinsic rewards from the geometric properties of the latent space, as shown in Figure \ref{fig:intro}. At its core, the \textbf{Iterative Robust Centroid Estimation (IRCE)} algorithm identifies a ``truth centroid'' from the \textbf{last hidden states} of terminal tokens, using their geometric relationship as a continuous reward. 
Unlike rule-based verifiers, our approach yields dense reward signals, which allows for more granular optimization across a wider range of reasoning scenarios where clear rules are unavailable. 
Compared to LLM-as-judge, Latent-GRPO eliminates external model dependencies, which significantly reduces training latency and prevents model collapse caused by inconsistent or noisy judging. 
More importantly, this approach effectively activates the rich reasoning knowledge acquired by the model during its large-scale pre-training phase. 
Detailed experiments on GSM8K, MATH, and Open-Platypus all show that Latent-GRPO achieves
2× training speedup compared to LLM-as-Judge across three model scales (0.6B, 1.7B, and 4B). While it exceeds accuracy of both LLM-as-Judge and Rule-based Methods.
Furthermore, additional analyses underscore the robustness and generalization ability of our approach across diverse scenarios.

\section{Related Work}
\noindent\textbf{Policy Optimization and Group-based Variants.}
Recent advances in reinforcement learning for LLMs have focused on improving training efficiency and stability. PPO~\cite{schulman2017proximal} was the foundational algorithm for RLHF~\cite{ouyang2022training,ziegler2019fine,stiennon2020learning}, but requires maintaining a critic model with significant memory overhead. DPO~\cite{rafailov2023direct} eliminated the critic but limits exploration capabilities. GRPO~\cite{shao2024deepseekmath} introduced group-based advantage estimation to balance online exploration and computational efficiency. Recent variants~\cite{yu2025dapo,zheng2025group,liu2025understanding,zhao2025geometric} address specific challenges at the optimizer level.

\noindent\textbf{Training-free evaluation Methods.}
To reduce dependence on expensive external supervision, researchers have explored training-free evaluation approaches. Methods like Self-Consistency~\cite{wang2022self}, Self-Refine~\cite{madaan2023self}, Tree-of-Thoughts~\cite{yao2023tree}, Best-of-N~\cite{stiennon2020learning}, and Forest-of-Thoughts~\cite{bi2024forest} leverage consensus mechanisms to identify high-quality outputs. 

\noindent\textbf{Latent Space and Latent Thinking.}
The latent space of LLMs is known to encode rich semantic information~\cite{zhang2023planner,goyal2023think,hao2024training,geiping2025scaling}. Recent work has explored latent thinking, using hidden states for self-evaluation and reward prediction. Methods like CoE~\cite{wang2024latent}, LTO~\cite{du2025latent}, LaTRO~\cite{chen2024language}, and EndoRM~\cite{li2025generalist} leverage latent representations for RL guidance. \textit{For comprehensive related work discussion, please refer to Appendix~\ref{app:related}}.

\section{Geometric Properties of Latent Space}
\label{sec:analysis}
This section demonstrates that the LLM latent space intrinsically captures reasoning quality through a series of empirical analyses. Specifically, we observe that correct reasoning trajectories exhibit high geometric clustering in their terminal hidden states, whereas incorrect paths remain scattered. Building on this, we explore leveraging these geometric features to score reasoning quality, ultimately finding a high degree of consistency with evaluations from external model-based verifiers. This alignment confirms the latent space as a robust and autonomous source of reward signals.

\begin{figure}[t]
\centering
\includegraphics[width=\columnwidth]{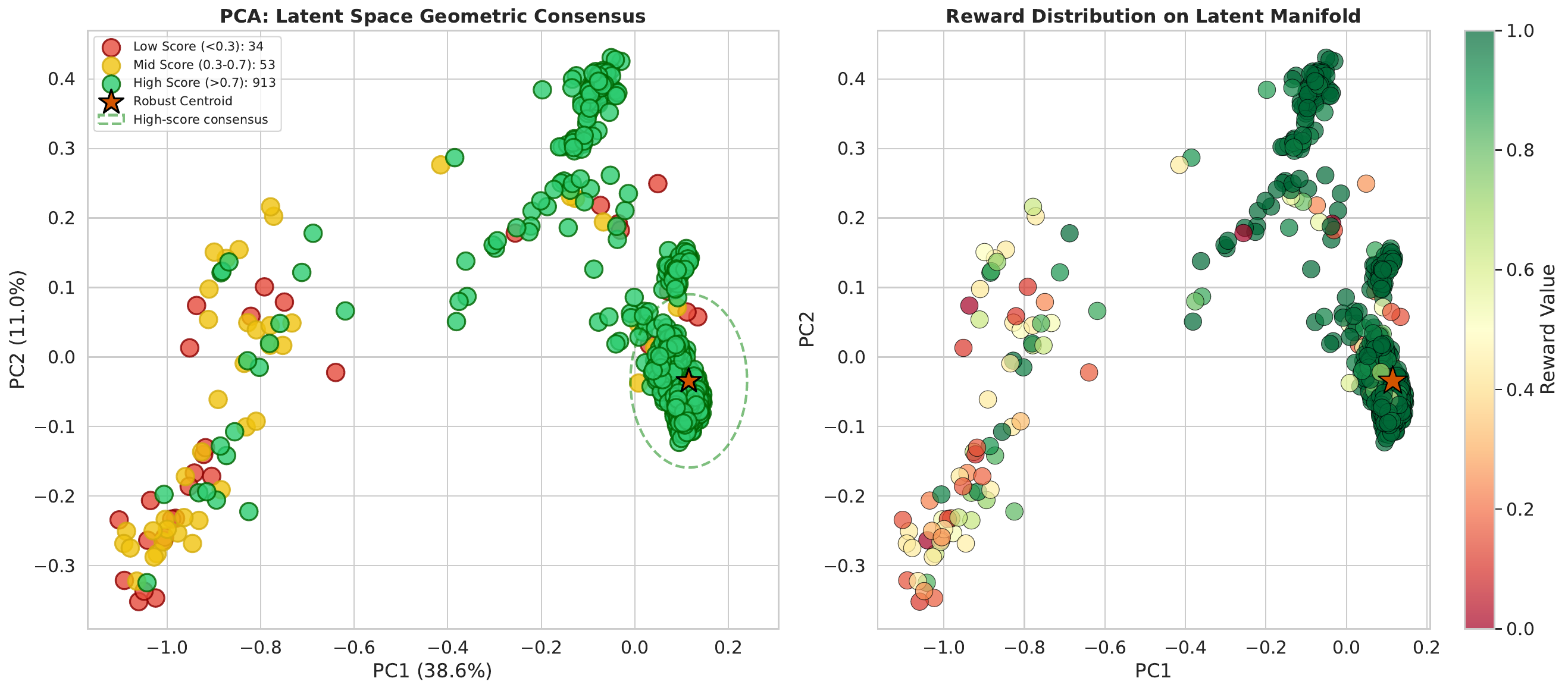}
\caption{2D PCA projection of 1,000 rollouts. Correct trajectories (green) form a dense consensus core around the truth centroid (gold star), while incorrect ones (red) scatter as outliers.}
\label{fig:clustering}
\end{figure}

\begin{figure*}[t]
\centering
\includegraphics[width=\textwidth]{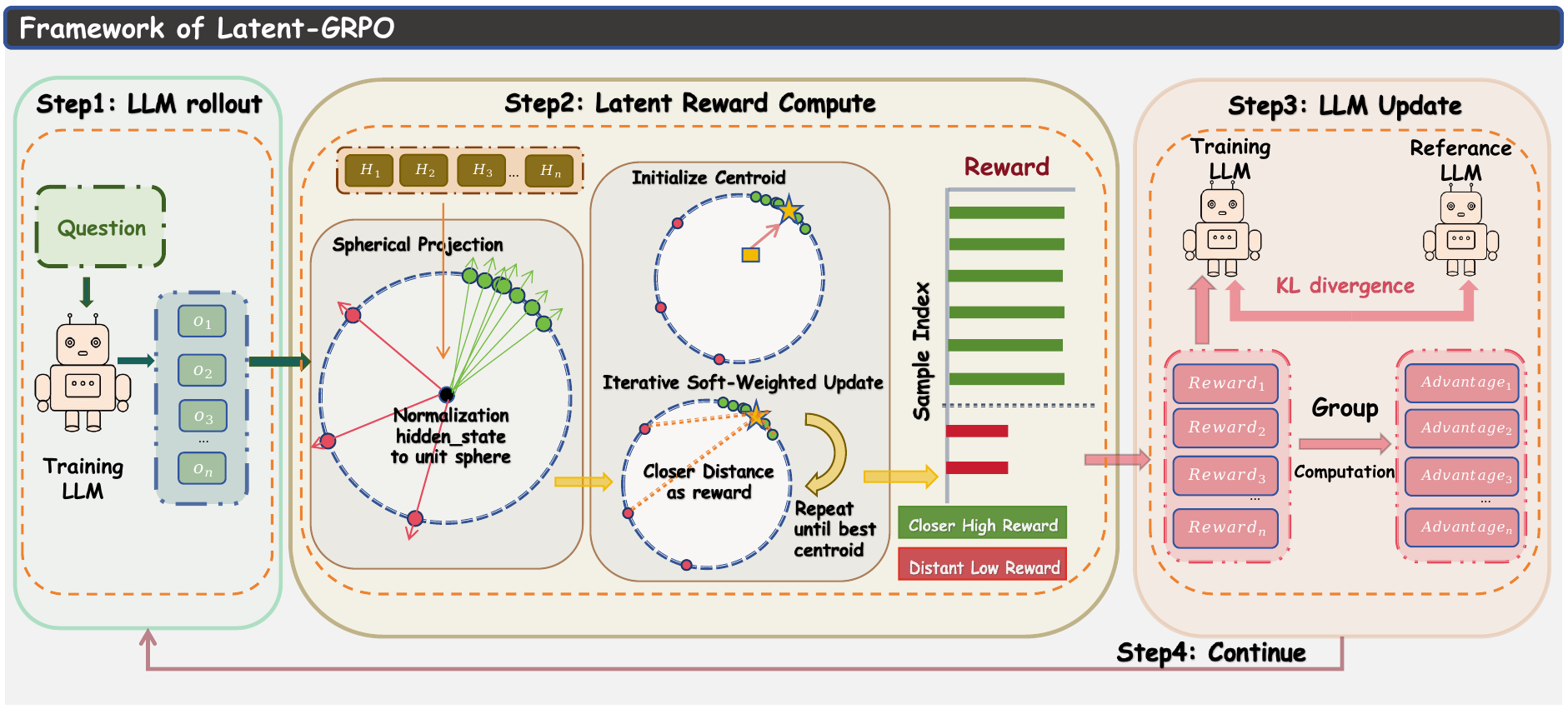}
\caption{Overview of the Latent-GRPO framework. The policy model generates a group of responses for each prompt. Instead of relying on external verifiers, we extract the hidden states of the last token from each trajectory and apply the Iterative Robust Centroid Estimation (IRCE) algorithm to compute intrinsic rewards based on geometric clustering in the latent space. These rewards are then used to compute group-relative advantages for policy optimization. The entire process operates within the latent space, achieving zero additional inference overhead while providing dense reward signals.}
\label{fig:framework}
\end{figure*}

\subsection{Theoretical Motivation}

Our analysis is guided by two fundamental properties of the Transformer architecture. First, the last hidden state of the terminal token, $h_T$, acts as a semantic summary of the entire reasoning chain \cite{afzal2025knowing, wang2024latent}. Specifically, as the final representation before the language model head, $h_T$ effectively aggregates the model's converged reasoning information \cite{shai2024transformers}. Second, successful reasoning often leads to semantic collapse \cite{papyan2020prevalence, wang2022self}. This means that while intermediate steps may vary, all correct trajectories tend to converge toward a unified semantic endpoint. In contrast, incorrect paths typically scatter due to their diverse failure modes \cite{marks2023geometry, zou2023representation}.

\subsection{Analysis Design}
To validate our findings, we conduct two analyses on the GSM8K dataset. First, we  generate 1,000 independent trajectories per prompt to examine global clustering patterns. Second, we construct GRPO groups ($G=8$) to evaluate the effectiveness of geometric reward assignment. In both cases, we extract the last hidden state of the terminal token, $h_{T} \in \mathbb{R}^{1024}$, and compute distances directly in the original high-dimensional space to preserve the full semantic information. To verify these geometric rewards, we use GPT-4o to obtain ground-truth labels; critically, these labels are used only for validation in this section and are not required by our training framework. 

\subsection{Large-Scale Clustering Analysis}

Figure~\ref{fig:clustering} visualizes the \textbf{last hidden states} of terminal tokens via PCA, revealing a clear ``core-periphery'' structure. Analyzing 1,000 trajectories from GSM8K, we observe two key patterns: (1) \textbf{High-Density Consensus Core}, where 913 correct trajectories cluster tightly around the centroid ($d_{\text{correct}} = 0.249$); and (2) \textbf{Dispersed Outlier Region}, where 34 incorrect trajectories remain scattered ($d_{\text{incorrect}} = 1.029$). The resulting 4.13$\times$ distance ratio confirms significant geometric separability of reasoning quality based on these \textbf{last hidden states}. To demonstrate the universality of these patterns, we provide additional observations across model scales (0.6B to 4B) and diverse datasets (e.g., ScienceQA and ARB) in Appendix~\ref{app:Geometric}.

\begin{figure}[t]
\centering
\includegraphics[width=\columnwidth]{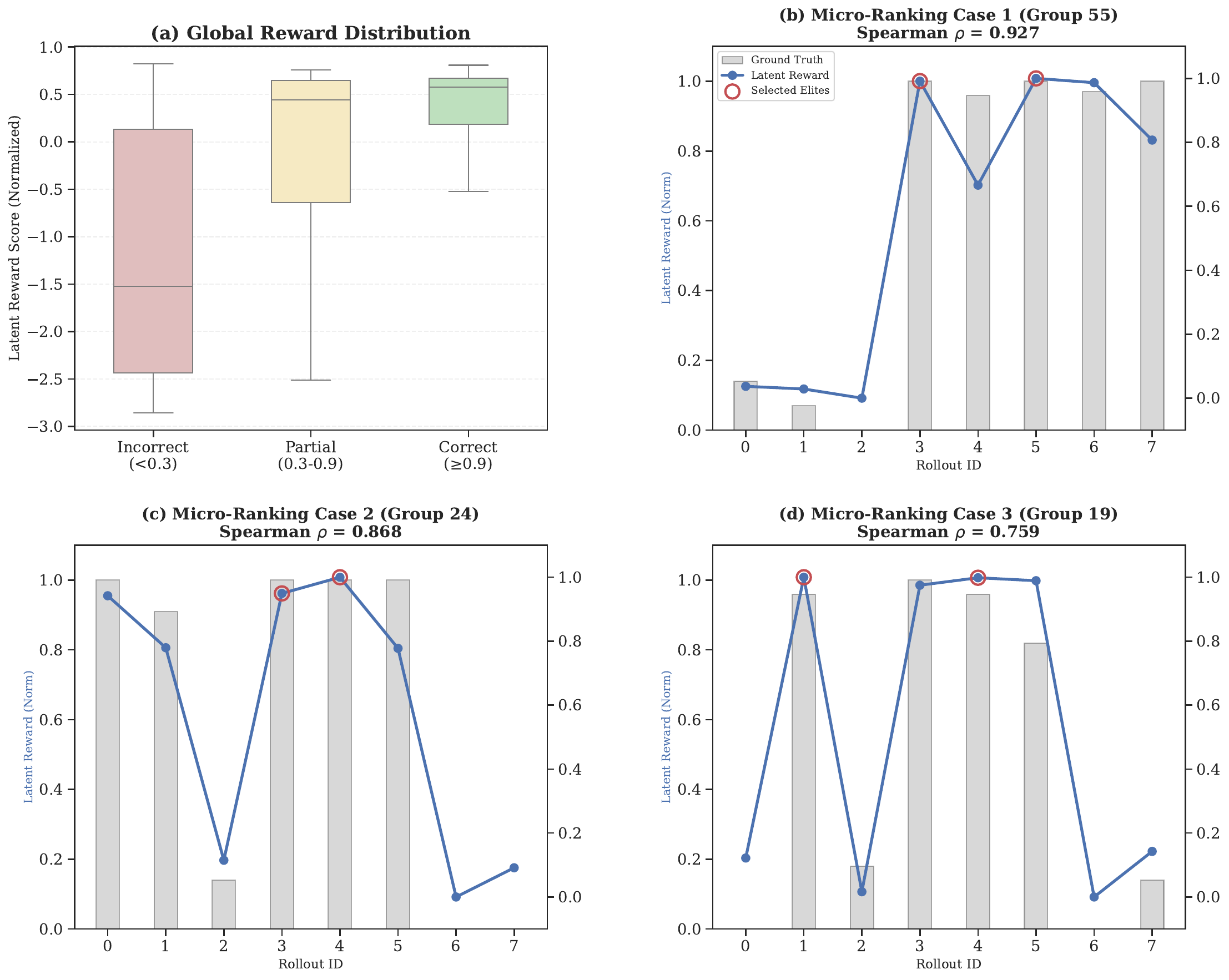}
\caption{Validation of geometric-based scoring. (a) Distribution separability across quality levels (correct, partial, incorrect). (b-d) Group-level ranking consistency in representative 8-trajectory groups.}
\label{fig:correlation}
\end{figure}

\subsection{Validation via GRPO Group Simulation}
Within each GRPO group, we calculate a robust centroid $\mathbf{c}$ via weighted aggregation of the terminal tokens' \textbf{last hidden states}. The latent geometric score is then defined as $s_{\text{latent}}(i) = -\|\mathbf{h}_i - \mathbf{c}\|_2$. Figure~\ref{fig:correlation} validates this approach: (a) presents a box plot of the reward distribution across different quality levels. It reveals that incorrect trajectories exhibit high dispersion and a lower median, whereas correct trajectories are significantly more concentrated with much higher reward values. (b-d) demonstrate high ranking consistency in representative groups. Notably, our method achieves a maximum Spearman correlation of $\rho=0.927$ and a Top-1 selection agreement of up to 85\% with external verifiers. These findings confirm that latent geometric properties can serve as a robust, training-free alternative to external judges, providing a solid foundation for Latent-GRPO. For a more detailed analysis, please refer to Appendix~\ref{app:section3ana}.

Altogether, these analysis demonstrate that the last hidden states of terminal tokens inherently capture reasoning quality through their geometric structure. This latent consistency provides a strong foundation for designing efficient reward mechanisms independent of external supervision. Building on these insights, we introduce the Latent-GRPO framework, which transforms these geometric properties into dense, continuous reward signals for policy optimization.

\section{Methodology: Latent-GRPO}
\label{sec:method}

\subsection{GRPO Short Review}
Group Relative Policy Optimization (GRPO) \cite{shao2024deepseekmath} optimizes policy performance by computing relative advantages across a group of $G$ trajectories $\{y_1, \dots, y_G\}$ sampled from the same prompt $x$:
\begin{equation}\small
A_i = \frac{R_i - \bar{R}_{\text{group}}}{\text{std}(R_{\text{group}}) + \epsilon}
\end{equation}
where $R_i$ represents the reward for the $i$-th trajectory. While architecturally efficient, standard GRPO faces two major challenges: heavy reliance on external verifiers, which leads to high costs and latency, and reward sparsity, which offers limited guidance for complex reasoning tasks. To address these issues, we propose a reward mechanism characterized by two key properties: (1) \textbf{intrinsic}, utilizing the model's own \textbf{last hidden states of terminal tokens}, and (2) \textbf{dense}, providing continuous quality scores that enhance optimization gradients.

\subsection{Core Algorithm: Iterative Robust Centroid Estimation}

Based on the geometric findings in Section~\ref{sec:analysis}, we introduce the Iterative Robust Centroid Estimation (IRCE) algorithm to extract high-fidelity reward signals. For each prompt, the policy model generates a group of $G$ trajectories. We first extract the \textbf{last hidden states of terminal tokens} $\mathbf{h}_i \in \mathbb{R}^d$ ($i=1, \dots, G$) and perform spherical normalization:

\begin{equation}\small
\tilde{\mathbf{h}}_i = \frac{\mathbf{h}_i}{\|\mathbf{h}_i\|_2}
\end{equation}
This operation projects all trajectories onto a unit hypersphere, effectively eliminating magnitude fluctuations while ensuring the subsequent analysis focuses exclusively on semantic directionality.

The core of IRCE is to dynamically estimate a consensus centroid $\boldsymbol{\mu}$ that represents the ``truth direction'' of reasoning within the group. To suppress the influence of erroneous outliers, we employ an iterative soft-weighting mechanism. In each iteration $s$, we compute soft weights $w_i^{(s)}$ using a Gaussian kernel based on the sample's current distance to the centroid:

\begin{equation}\small
w_i^{(s)} = \frac{\exp\left(-(d_i^{(s)})^2 / 2(\sigma^{(s)})^2\right)}{\sum_{j=1}^G \exp\left(-(d_j^{(s)})^2 / 2(\sigma^{(s)})^2\right)}
\end{equation}

where $\sigma^{(s)}$ is an adaptive scale parameter derived from the group's distance distribution. The centroid is subsequently updated via weighted aggregation: $\boldsymbol{\mu}^{(s+1)} = \text{Norm}(\sum w_i^{(s)} \tilde{\mathbf{h}}_i)$. \textbf{Upon convergence} after $T$ iterations, the intrinsic reward $R_i$ is defined as the negative Euclidean distance to the final centroid:

\begin{equation}\small
R_i = \frac{-\|\tilde{\mathbf{h}}_i - \boldsymbol{\mu}^{(T)}\|_2 - \min_j(-\|\tilde{\mathbf{h}}_j - \boldsymbol{\mu}^{(T)}\|_2)}{\max_j(-\|\tilde{\mathbf{h}}_j - \boldsymbol{\mu}^{(T)}\|_2) - \min_j(-\|\tilde{\mathbf{h}}_j - \boldsymbol{\mu}^{(T)}\|_2)}
\end{equation}
This mechanism yields intrinsic rewards that are \textbf{dense, continuous, and naturally bounded}. By applying Min-Max normalization, we map the distances into a calibrated range $R_i \in [0, 1]$, which prevents gradient explosion and ensures optimization stability across different reasoning tasks. The complete procedure, including the adaptive scaling and normalization, is summarized in Algorithm~\ref{alg:irce}. Furthermore, we provide a rigorous mathematical derivation of this iterative weighting scheme in Appendix~\ref{app:irce} to demonstrate its robustness against sampling noise.

\begin{algorithm}[t]\small
\caption{Iterative Robust Centroid Estimation}
\label{alg:irce}
\begin{algorithmic}[1]
\Require Hidden states $\{\mathbf{h}_1, ..., \mathbf{h}_G\}$, max iterations $T$
\Ensure Rewards $\{R_1, ..., R_G\}$
\State \textbf{// Step 1: Spherical Projection}
\For{$i = 1$ to $G$}
    \State $\tilde{\mathbf{h}}_i \Leftarrow \mathbf{h}_i / \|\mathbf{h}_i\|_2$
\EndFor
\State \textbf{// Step 2: Initialize Centroid}
\State $\boldsymbol{\mu} \Leftarrow \frac{1}{G} \sum_{i=1}^{G} \tilde{\mathbf{h}}_i$
\State $\boldsymbol{\mu} \Leftarrow \boldsymbol{\mu} / \|\boldsymbol{\mu}\|_2$
\State \textbf{// Step 3: Iterative Soft-Weighted Update}
\For{$s = 0$ to $T-1$}
    \For{$i = 1$ to $G$}
        \State $d_i \Leftarrow \|\tilde{\mathbf{h}}_i - \boldsymbol{\mu}\|_2$
    \EndFor
    \State $\sigma \Leftarrow \text{std}(\{d_1, ..., d_G\}) + \epsilon$
    \For{$i = 1$ to $G$}
        \State $w_i \Leftarrow \exp(-(d_i)^2 / (2\sigma^2))$
    \EndFor
    \State $\mathbf{w} \Leftarrow \mathbf{w} / \sum_j w_j$ \Comment{Normalize weights}
    \State $\boldsymbol{\mu} \Leftarrow \sum_i w_i \tilde{\mathbf{h}}_i$
    \State $\boldsymbol{\mu} \Leftarrow \boldsymbol{\mu} / \|\boldsymbol{\mu}\|_2$ \Comment{Normalize centroid}
\EndFor
\State \textbf{// Step 4: Compute Rewards}
\For{$i = 1$ to $G$}
    \State $d_i \Leftarrow \|\tilde{\mathbf{h}}_i - \boldsymbol{\mu}\|_2$
    \State $R_i \Leftarrow -d_i$
\EndFor
\State $\mathbf{R} \Leftarrow \text{MinMaxNormalize}(\mathbf{R})$
\State \Return $\mathbf{R}$
\end{algorithmic}
\end{algorithm}

\subsection{Latent-GRPO Framework}
As illustrated in Figure~\ref{fig:framework}, \textbf{Latent-GRPO} integrates the IRCE algorithm into the GRPO pipeline. Unlike traditional reinforcement learning which relies on static external verifiers, Latent-GRPO adopts a dynamic, intrinsic reward mechanism that operates without external supervision.

\begin{table*}[h!]
\small
\centering
\begin{tabular*}{\textwidth}{@{\extracolsep{\fill}}cccccccc@{}}
\toprule
\multirow{2}{*}{\textbf{Dataset}} & \multirow{2}{*}{\textbf{Method}} & \multicolumn{2}{c}{\textbf{Qwen3-0.6B}} & \multicolumn{2}{c}{\textbf{Qwen3-1.7B}} & \multicolumn{2}{c}{\textbf{Qwen3-4B}} \\
\cmidrule{3-4} \cmidrule{5-6} \cmidrule{7-8}
 &  & \textbf{Acc} $\uparrow$ & \textbf{Time} $\downarrow$ & \textbf{Acc} $\uparrow$ & \textbf{Time} $\downarrow$ & \textbf{Acc} $\uparrow$ & \textbf{Time} $\downarrow$ \\
\midrule
\multirow{3}{*}{\textit{\textbf{GSM8K}}} & LLM-as-Judge & 53.52\% & 768.42m & 64.20\% & 1032.55m & 72.12\% & 1411.72m \\
 & Rule-based & 58.41\% & 434.61m & 71.55\% & \textbf{488.63m} & 79.87\% & \textbf{651.45m} \\
 & \textbf{Latent-GRPO (Ours)}& \textbf{61.25\%} & \textbf{431.18m} & \textbf{73.88\%} & 492.34m & \textbf{82.34\%} & 658.21m \\
\midrule
\multirow{3}{*}{\textit{\textbf{MATH}}} & LLM-as-Judge &52.94\% & 1224.15m & 65.77\% & 1608.34m & 77.44\% & 2357.31m \\
 & Rule-based & 55.63\% & 723.12m & 42.14\% & 814.22m & 62.63\% & 1084.72m \\
 & \textbf{Latent-GRPO (Ours)} & \textbf{58.47\%} & \textbf{718.63m} & \textbf{78.51\%} & \textbf{811.51m} & \textbf{77.53\%} & \textbf{1081.47m} \\
\midrule
\multirow{2}{*}{\textit{\textbf{Open-Platypus}}} & LLM-as-Judge & 34.45\% & 1937.82m & 56.69\% & 2573.41m & 65.21\% & 3522.18m \\
 & \textbf{Latent-GRPO (Ours)}& \textbf{40.56\%} & \textbf{1079.27m} & \textbf{64.82\%} & \textbf{1218.92m} & \textbf{78.06\%} & \textbf{1632.52m} \\
\bottomrule
\end{tabular*}
\caption{\label{tab:main_comparison}Comprehensive comparison of reward methods across datasets and model scales. The results demonstrate that Latent-GRPO consistently achieves superior accuracy and training efficiency (Time per epoch) compared to LLM-as-Judge and Rule-based baselines. (Simulated QPS for LLM-as-Judge is 2)}
\end{table*}

\textbf{Dynamic Adaptability.} As the policy model evolves during training, its latent representation space naturally shifts. Unlike static external verifiers, IRCE dynamically estimates the consensus centroid for each training batch, allowing reward signals to adapt in real-time to the model's current state. This online adaptation effectively mitigates distribution shift—a common challenge when reward signals become misaligned with evolving model representations. Additionally, the iterative soft-weighting mechanism ensures stable and robust gradient signals even when initial sample quality is suboptimal.

\textbf{Computational Efficiency.} Latent-GRPO eliminates the computational overhead of external verification by leveraging hidden states already computed during rollout. While conventional verifiers require additional forward passes scaling as $O(GL)$ (where $G$ is group size and $L$ is sequence length), IRCE operates on the last hidden states of terminal tokens with complexity $O(GTd)$ (where $T$ is the number of iterations and $d$ is the latent dimension). Since $d$ is fixed by model architecture and $T \ll L$, the overhead becomes negligible. By eliminating the need for separate reward models or value functions, Latent-GRPO simultaneously reduces memory footprint and transforms reward computation from an external bottleneck into an efficient intrinsic process.

\section{Experimental Setup}
\label{sec:setup}
This section describes our experimental configuration, including dataset selection, evaluation metrics, hardware setup, and the design of our main experiment and ablation studies.

\subsection{Datasets and Evaluation Metrics}
\noindent\textbf{Datasets.} We evaluate our approach on three complementary datasets that span different reasoning complexity levels and domains. \textbf{GSM8K} \cite{cobbe2021training} contains elementary-level mathematical word problems, establishing a foundation for basic arithmetic reasoning. \textbf{MATH} \cite{hendrycks2021measuring} comprises high school and competition-level problems, testing more sophisticated mathematical reasoning. \textbf{Open-Platypus} \cite{lee2023platypus} covers diverse reasoning tasks across physics, logic, and mathematics, enabling validation across multiple domains. Together, these datasets provide comprehensive coverage for evaluating both accuracy improvements and computational efficiency across varying reasoning complexities. The spilt of train/test datasets and detailed dataset statistics are provided in Appendix~\ref{app:datasets}.

\noindent\textbf{Evaluation Metrics.} We measure two primary performance dimensions: (1) \textbf{Task Accuracy} on each test dataset of benchmark to assess reasoning capability, and (2) \textbf{Training Efficiency} measured as time per epoch across different reward methods under identical experimental settings.

\subsection{Experimental Setup}
\noindent\textbf{Models and Hardware.} We validate our method across three Qwen models of varying sizes: Qwen3-0.6B, Qwen3-1.7B, and Qwen3-4B. This range enables assessment of our method's effectiveness across different model scales. All experiments are conducted on a single GPU. For detailed hyperparameter settings including GRPO training configuration, IRCE algorithm parameters, and hardware specifications, please refer to Appendix~\ref{app:hyperparameters}.

\noindent\textbf{Main Experiment} We compare three reward paradigms across three datasets (GSM8K, MATH, Open-Platypus) and three model sizes (Qwen3-0.6B, Qwen3-1.7B, Qwen3-4B): (1) \textbf{LLM-as-Judge} using external verification (GPT-4o), (2) \textbf{Rule-based} verification with ground-truth labels, and (3) our proposed \textbf{Latent-GRPO} leveraging intrinsic latent space geometry. This comparison evaluates whether intrinsic geometric signals can match or exceed external verification in both accuracy and training efficiency. Detailed baseline descriptions are provided in Appendix~\ref{app:basemethod}.

\begin{table}[t]
\small
\centering
\begin{tabularx}{\columnwidth}{Xcc} 
\toprule
\textbf{Method} & \textbf{Acc} $\uparrow$ & \textbf{Time (m)} $\downarrow$ \\
\midrule
\rowcolor{lightgray} \multicolumn{3}{l}{\textit{\textbf{Qwen3-0.6B}}} \\
Mean Pooling & 58.74\%          & 435.22          \\
Weighted Mean & 57.12\%          & 442.89          \\
\rowcolor{blue!10}
\textbf{Last Token (Ours)} & \textbf{61.25\%} & \textbf{431.18} \\
\midrule
\rowcolor{lightgray} \multicolumn{3}{l}{\textit{\textbf{Qwen3-1.7B}}} \\
Mean Pooling & 71.05\%          & 497.61          \\
Weighted Mean & 69.88\%          & 512.14          \\
\rowcolor{blue!10}
\textbf{Last Token (Ours)} & \textbf{73.88\%} & \textbf{492.34} \\
\midrule
\rowcolor{lightgray} \multicolumn{3}{l}{\textit{\textbf{Qwen3-4B}}} \\
Mean Pooling & 79.45\%          & 664.33          \\
Weighted Mean & 78.12\%          & 685.56          \\
\rowcolor{blue!10}
\textbf{Last Token (Ours)} & \textbf{82.34\%} & \textbf{658.21} \\
\bottomrule
\end{tabularx}
\caption{\label{tab:extraction} Comparison of different hidden state extraction methods. Using the terminal token's representation (Last Token) consistently yields the best reasoning performance with the lowest computational latency.}
\end{table}

\section{Results and Analysis}
\label{sec:results}
This section presents a comprehensive evaluation of \textbf{Latent-GRPO} across three dimensions: performance and efficiency compared to external verifiers, core design (IRCE) choices through ablation studies, and generalization across model families and unseen tasks.

\subsection{Reward Methods Comparison}
We compare three reward paradigms on GSM8K and MATH: LLM-as-Judge (GPT-4o), Rule-based verification, and Latent-GRPO. For Open-Platypus, we focus on LLM-as-Judge and Latent-GRPO since rule-based methods are limited to tasks with deterministic ground truth.

As shown in Label \ref{tab:main_comparison}, Latent-GRPO achieves approximately \textbf{2× training speedup} compared to LLM-as-Judge across all datasets and model scales. Meanwhile, Latent-GRPO maintains or exceeds the accuracy of both LLM-as-Judge and Rule-based verification. Detailed per-dataset and per-model results are provided in Appendix~\ref{app:main}. 

\noindent\textbf{Efficiency.} Latent-GRPO eliminates the external verifier bottleneck that constrains LLM-as-Judge. The latter faces two system-level costs: API rate limiting (2 QPS) forces scoring requests into a queue, and each API call incurs 1-2 minutes of latency. Together, these consume 58-63\% of total training time. Latent-GRPO instead computes rewards on hidden states already available from the forward pass using the IRCE algorithm, requiring only geometric operations with $O(GTd)$ complexity (where $T \ll L$, typically 5 iterations vs. 2048 sequence length), compared to $O(GL)$ for transformer-based verifiers. Rule-based verification operates at similar cost to Latent-GRPO, confirming that speedup comes primarily from eliminating external calls.

\noindent\textbf{Accuracy.} This advantage stems from two factors. First, Latent-GRPO provides continuous, dense rewards based on distance to the consensus centroid, whereas Rule-based and LLM-as-Judge provide only binary 0/1 feedback. Richer reward signals enable more effective policy optimization. Second, Latent-GRPO derives rewards from the model's internal geometry rather than external judges, avoiding inconsistency and noise from external verifiers. The model no longer depends on external verification accuracy, which stabilizes training and prevents collapse.

\begin{table}[t]
\centering
\small
\begin{tabularx}{\columnwidth}{Xcc}
\toprule
\textbf{Method} & \textbf{Acc} $\uparrow$ & \textbf{Time (m)} $\downarrow$ \\
\midrule
\rowcolor{lightgray} \multicolumn{3}{l}{\textit{\textbf{Qwen3-0.6B}}} \\
Mean Pool        & 57.12\%          & 452.34          \\
K-Means          & 58.85\%          & 489.12          \\
Eigen Centrality & 59.43\%          & 468.76          \\
\rowcolor{blue!10}
\textbf{IRCE (Ours)} & \textbf{61.25\%} & \textbf{431.18} \\
\midrule
\rowcolor{lightgray} \multicolumn{3}{l}{\textit{\textbf{Qwen3-1.7B}}} \\
Mean Pool        & 68.45\%          & 512.67          \\
K-Means          & 70.12\%          & 543.89          \\
Eigen Centrality & 71.56\%          & 531.42          \\
\rowcolor{blue!10}
\textbf{IRCE (Ours)} & \textbf{73.88\%} & \textbf{492.34} \\
\midrule
\rowcolor{lightgray} \multicolumn{3}{l}{\textit{\textbf{Qwen3-4B}}} \\
Mean Pool        & 77.89\%          & 682.12          \\
K-Means          & 79.23\%          & 725.67          \\
Eigen Centrality & 80.56\%          & 708.45          \\
\rowcolor{blue!10}
\textbf{IRCE (Ours)} & \textbf{82.34\%} & \textbf{658.21} \\
\bottomrule
\end{tabularx}
\caption{\label{tab:latent_score}Comparison of different latent consensus scoring methods on GSM8K. IRCE (Ours) achieves the highest reasoning accuracy with the minimum computational overhead across all model scales.}
\end{table}

\begin{table*}[t]
\small
\centering
\begin{tabularx}{\textwidth}{Xccccccc}
\toprule
\textbf{Method} & \textbf{AIME24} $\uparrow$ & \textbf{AIME25} $\uparrow$ & \textbf{MATH-500} $\uparrow$ & \textbf{MMLU} $\uparrow$ & \textbf{BBH} $\uparrow$ & \textbf{Avg} $\uparrow$ & \textbf{Time (m) $\downarrow$} \\
\midrule
\rowcolor{lightgray} \multicolumn{8}{l}{\textit{\textbf{Qwen3-0.6B}}} \\
Base                        & 10.7          & 15.1          & 77.6          & 52.8          & 41.5          & 39.54          & –          \\
GRPO (LLM-Judge)            & 9.4           & 11.7          & 65.9          & 50.6          & 42.4          & 36.00          & 4082.96          \\
\rowcolor{blue!10}
\textbf{Latent-GRPO (Ours)} & \textbf{10.6} & \textbf{19.2} & \textbf{67.3} & \textbf{49.9} & \textbf{39.2} & \textbf{37.24} & 2280.52          \\
\midrule
\rowcolor{lightgray} \multicolumn{8}{l}{\textit{\textbf{Qwen3-1.7B}}} \\
Base                        & 48.3          & 36.8          & 93.4          & 62.6          & 54.5          & 59.12          & –          \\
GRPO (LLM-Judge)            & 48.1          & 33.8          & 90.6          & 67.6          & 67.4          & 61.50          & 4988.84         \\
\rowcolor{blue!10}
\textbf{Latent-GRPO (Ours)} & \textbf{44.6} & \textbf{35.4} & \textbf{91.2} & \textbf{68.1} & \textbf{68.7} & \textbf{61.60} & 2340.05          \\
\midrule
\rowcolor{lightgray} \multicolumn{8}{l}{\textit{\textbf{Qwen3-4B}}} \\
Base                        & 73.8          & 65.6          & 97.0          & 83.7          & 72.6          & 78.54          & –          \\
GRPO (LLM-Judge)            & 72.9          & 63.3          & 96.4          & 85.5          & 81.9          & 80.00          & 6753.47          \\
\rowcolor{blue!10}
\textbf{Latent-GRPO (Ours)} & \textbf{74.6} & \textbf{66.7} & \textbf{97.5} & \textbf{88.5} & \textbf{82.3} & \textbf{81.92} & 3108.44          \\
\bottomrule
\end{tabularx}
\caption{\label{tab:benchmarks2}Performance and efficiency comparison across Qwen3 scales on mathematics (AIME, MATH-500) and general reasoning (MMLU, BBH) benchmarks. }
\end{table*}

\subsection{Ablation Studies: Core Design Choices}
We validate two critical design choices through systematic ablation studies: (1) The hidden state extraction method for capturing reasoning quality, and (2) The centroid estimation algorithm for computing intrinsic rewards. Detailed quantitative analysis and per-model comparisons are provided in Appendix~\ref{app:abl}.

\noindent\textbf{Hidden State Extraction.} We compare three approaches for extracting reasoning quality from hidden states: \textbf{(1) Last Token}, using only the terminal token's representation; \textbf{(2) Mean Pooling}, averaging representations across all tokens; \textbf{(3) Weighted Mean}, computing weighted averages of key tokens (weighted averaging of pre-designed key tokens such as mathematical operators, structural markers, and reasoning keywords).

As illustrated in Table \ref{tab:extraction}, the Last Token method consistently outperforms all aggregation baselines across various model scales. This finding reveals a fundamental property of transformer-based reasoning: the final token, as the immediate precursor to the end-of-sequence (EOS) prediction, acts as a semantic bottleneck where reasoning correctness is crystallized. In contrast, Mean Pooling degrades performance by incorporating noise from intermediate tokens that are often orthogonal to the final correctness. The failure of Weighted Mean, which focuses on linguistically significant keywords (e.g., "therefore", "solution") is particularly instructive. It suggests that reasoning quality is not localized at specific lexical markers but emerges holistically through the generation process, ultimately converging into the final representation.

\noindent\textbf{Centroid Estimation} We compare four approaches for estimating the consensus centroid from a group of trajectories: \textbf{(1) Mean Pooling},  \textbf{(2) K-Means clustering}, \textbf{(3) Eigen Centrality}, and  \textbf{(4) IRCE}, our proposed iterative robust centroid estimation. All experiments are conducted on GSM8K across three model scales using the Last Token extraction method identified above. For details of these latent reward methods, please refer to Appendix~\ref{app:latent_reward}. 

As shown in Table~\ref{tab:latent_score}, IRCE consistently outperforms all baseline methods across all model scales. Mean Pooling fails to handle outliers effectively, resulting in suboptimal centroids. K-Means attempts to improve through hard cluster assignment and achieves competitive accuracy, yet still underperforms IRCE. Eigen Centrality uses graph-based importance weighting but introduces significant computational overhead through eigendecomposition, achieving lower accuracy while incurring higher computational cost. Our method, IRCE, maintains both robustness and efficiency, making it the optimal choice for latent reward estimation.

\subsection{Generalization}

\begin{table}[t]
\small
\centering
\begin{tabularx}{\columnwidth}{Xcc}
\toprule
\textbf{Method} & \textbf{Acc} $\uparrow$ & \textbf{Time (m)} $\downarrow$ \\
\midrule
\rowcolor{lightgray} \multicolumn{3}{l}{\textit{\textbf{GSM8K (2k steps)}}} \\
LLM-as-Judge & 71.34\%          & 1284.56          \\
\rowcolor{blue!10}
\textbf{Latent-GRPO}  & \textbf{78.62\%} & \textbf{591.24} \\
\midrule
\rowcolor{lightgray} \multicolumn{3}{l}{\textit{\textbf{MATH (3k steps)}}} \\
LLM-as-Judge & 43.12\%          & 1945.12          \\
\rowcolor{blue!10}
\textbf{Latent-GRPO}  & \textbf{52.45\%} & \textbf{912.87} \\
\midrule
\rowcolor{lightgray} \multicolumn{3}{l}{\textit{\textbf{Open-Platypus (4.5k steps)}}} \\
LLM-as-Judge & 61.88\%          & 2915.68          \\
\rowcolor{blue!10}
\textbf{Latent-GRPO}  & \textbf{73.12\%} & \textbf{1386.42} \\
\bottomrule
\end{tabularx}
\caption{\label{tab:llama} Experimental results of Latent-GRPO on Llama3.2-3B. Following the same protocol as Qwen3 experiments, Latent-GRPO demonstrates superior scaling performance and significantly reduced training latency compared to the LLM-as-judge baseline.}
\end{table}

\noindent\textbf{Capability Preservation.} 
A critical concern in RL training is the potential loss of general capabilities due to task-specific overfitting. 
To address this, we train Latent-GRPO on a specific mixture of reasoning datasets and subsequently evaluate it on a suite of unseen general benchmarks, including MMLU, AIME (24 \& 25), BBH, and MATH-500. 
As illustrated in Table~\ref{tab:benchmarks2}, Latent-GRPO consistently maintains or surpasses the performance of base models across these diverse tasks. 
These results confirm that dense intrinsic rewards guide the model toward mastering transferable reasoning patterns rather than overfitting.

\noindent\textbf{Cross-Model Generalization.} 
To validate that Latent-GRPO generalizes beyond the Qwen3 series, we extend our evaluation to Llama3.2-3B on the GSM8K, MATH, and Open-Platypus datasets. 
As shown in Table \ref{tab:llama}, the latent space geometry provides a universal signal for reasoning quality that is independent of the model family.

\section{Conclusion}
\label{sec:conclusion}
In this work, we introduce Latent-GRPO, a training framework designed to overcome the efficiency bottlenecks and sparse rewards inherent in verifier-dependent reinforcement learning. 
At the core of this framework lies the Iterative Robust Centroid Estimation algorithm, which transforms latent geometry into dense rewards, eliminating the reliance on slow external verifiers and achieving a substantial acceleration in training compared to the LLM-as-a-judge baseline.
Furthermore, extensive evaluations on unseen benchmarks confirm that our approach maintains competitive accuracy. 
It effectively preserves native capabilities and prevents task-specific overfitting. 
These findings substantiate that LLMs possess inherent self-evaluation mechanisms and offer a scalable paradigm for verifier-free post-training.

\section*{Limitations}
\label{sec:limitations}
Our work has two primary limitations that we aim to address in future research. First, while \textbf{Latent-GRPO} is effective up to 8B parameters, its scaling behavior in ultra-large models (70B+) and applicability to open-ended generation remain to be explored. Second, while we provide strong empirical evidence, a formal mathematical framework for latent clustering is still nascent. Moving forward, we plan to extend the geometric consensus hypothesis to broader tasks and investigate hybrid frameworks that integrate \textbf{IRCE} rewards with offline paradigms like DPO to further stabilize self-supervised alignment.

\bibliography{custom}

@article{schulman2017proximal,
  title={Proximal policy optimization algorithms},
  author={Schulman, John and Wolski, Filip and Dhariwal, Prafulla and Radford, Alec and Klimov, Oleg},
  journal={arXiv preprint arXiv:1707.06347},
  year={2017}
}

@article{rafailov2023direct,
  title={Direct preference optimization: Your language model is secretly a reward model},
  author={Rafailov, Rafael and Sharma, Archit and Mitchell, Eric and Manning, Christopher D and Ermon, Stefano and Finn, Chelsea},
  journal={Advances in neural information processing systems},
  volume={36},
  pages={53728--53741},
  year={2023}
}

@article{shao2024deepseekmath,
  title={Deepseekmath: Pushing the limits of mathematical reasoning in open language models},
  author={Shao, Zhihong and Wang, Peiyi and Zhu, Qihao and Xu, Runxin and Song, Junxiao and Bi, Xiao and Zhang, Haowei and Zhang, Mingchuan and Li, YK and Wu, Yang and others},
  journal={arXiv preprint arXiv:2402.03300},
  year={2024}
}

@article{yu2025dapo,
  title={Dapo: An open-source llm reinforcement learning system at scale},
  author={Yu, Qiying and Zhang, Zheng and Zhu, Ruofei and Yuan, Yufeng and Zuo, Xiaochen and Yue, Yu and Dai, Weinan and Fan, Tiantian and Liu, Gaohong and Liu, Lingjun and others},
  journal={arXiv preprint arXiv:2503.14476},
  year={2025}
}

@article{zheng2025group,
  title={Group sequence policy optimization},
  author={Zheng, Chujie and Liu, Shixuan and Li, Mingze and Chen, Xiong-Hui and Yu, Bowen and Gao, Chang and Dang, Kai and Liu, Yuqiong and Men, Rui and Yang, An and others},
  journal={arXiv preprint arXiv:2507.18071},
  year={2025}
}

@article{liu2025understanding,
  title={Understanding r1-zero-like training: A critical perspective},
  author={Liu, Zichen and Chen, Changyu and Li, Wenjun and Qi, Penghui and Pang, Tianyu and Du, Chao and Lee, Wee Sun and Lin, Min},
  journal={arXiv preprint arXiv:2503.20783},
  year={2025}
}

@article{zhao2025geometric,
  title={Geometric-mean policy optimization},
  author={Zhao, Yuzhong and Liu, Yue and Liu, Junpeng and Chen, Jingye and Wu, Xun and Hao, Yaru and Lv, Tengchao and Huang, Shaohan and Cui, Lei and Ye, Qixiang and others},
  journal={arXiv preprint arXiv:2507.20673},
  year={2025}
}

@article{jacobs1991adaptive,
  title={Adaptive mixtures of local experts},
  author={Jacobs, Robert A and Jordan, Michael I and Nowlan, Steven J and Hinton, Geoffrey E},
  journal={Neural computation},
  volume={3},
  number={1},
  pages={79--87},
  year={1991},
  publisher={MIT Press}
}

@inproceedings{du2022glam,
  title={Glam: Efficient scaling of language models with mixture-of-experts},
  author={Du, Nan and Huang, Yanping and Dai, Andrew M and Tong, Simon and Lepikhin, Dmitry and Xu, Yuanzhong and Krikun, Maxim and Zhou, Yanqi and Yu, Adams Wei and Firat, Orhan and others},
  booktitle={International conference on machine learning},
  pages={5547--5569},
  year={2022},
  organization={PMLR}
}

@article{lee2023rlaif,
  title={Rlaif: Scaling reinforcement learning from human feedback with ai feedback},
  author={Lee, Harrison and Phatale, Samrat and Mansoor, Hassan and Lu, Kellie Ren and Mesnard, Thomas and Ferret, Johan and Bishop, Colton and Hall, Ethan and Carbune, Victor and Rastogi, Abhinav},
  year={2023}
}

@article{ouyang2022training,
  title={Training language models to follow instructions with human feedback},
  author={Ouyang, Long and Wu, Jeffrey and Jiang, Xu and Almeida, Diogo and Wainwright, Carroll and Mishkin, Pamela and Zhang, Chong and Agarwal, Sandhini and Slama, Katarina and Ray, Alex and others},
  journal={Advances in neural information processing systems},
  volume={35},
  pages={27730--27744},
  year={2022}
}

@article{ziegler2019fine,
  title={Fine-tuning language models from human preferences},
  author={Ziegler, Daniel M and Stiennon, Nisan and Wu, Jeffrey and Brown, Tom B and Radford, Alec and Amodei, Dario and Christiano, Paul and Irving, Geoffrey},
  journal={arXiv preprint arXiv:1909.08593},
  year={2019}
}

@article{stiennon2020learning,
  title={Learning to summarize with human feedback},
  author={Stiennon, Nisan and Ouyang, Long and Wu, Jeffrey and Ziegler, Daniel and Lowe, Ryan and Voss, Chelsea and Radford, Alec and Amodei, Dario and Christiano, Paul F},
  journal={Advances in neural information processing systems},
  volume={33},
  pages={3008--3021},
  year={2020}
}

@article{wang2022self,
  title={Self-consistency improves chain of thought reasoning in language models},
  author={Wang, Xuezhi and Wei, Jason and Schuurmans, Dale and Le, Quoc and Chi, Ed and Narang, Sharan and Chowdhery, Aakanksha and Zhou, Denny},
  journal={arXiv preprint arXiv:2203.11171},
  year={2022}
}

@article{madaan2023self,
  title={Self-refine: Iterative refinement with self-feedback},
  author={Madaan, Aman and Tandon, Niket and Gupta, Prakhar and Hallinan, Skyler and Gao, Luyu and Wiegreffe, Sarah and Alon, Uri and Dziri, Nouha and Prabhumoye, Shrimai and Yang, Yiming and others},
  journal={Advances in Neural Information Processing Systems},
  volume={36},
  pages={46534--46594},
  year={2023}
}

@article{yao2023tree,
  title={Tree of thoughts: Deliberate problem solving with large language models},
  author={Yao, Shunyu and Yu, Dian and Zhao, Jeffrey and Shafran, Izhak and Griffiths, Tom and Cao, Yuan and Narasimhan, Karthik},
  journal={Advances in neural information processing systems},
  volume={36},
  pages={11809--11822},
  year={2023}
}

@article{bi2024forest,
  title={Forest-of-thought: Scaling test-time compute for enhancing llm reasoning},
  author={Bi, Zhenni and Han, Kai and Liu, Chuanjian and Tang, Yehui and Wang, Yunhe},
  journal={arXiv preprint arXiv:2412.09078},
  year={2024}
}

@article{sun2024fast,
  title={Fast best-of-n decoding via speculative rejection},
  author={Sun, Hanshi and Haider, Momin and Zhang, Ruiqi and Yang, Huitao and Qiu, Jiahao and Yin, Ming and Wang, Mengdi and Bartlett, Peter and Zanette, Andrea},
  journal={Advances in Neural Information Processing Systems},
  volume={37},
  pages={32630--32652},
  year={2024}
}

@article{wang2024latent,
  title={Latent space chain-of-embedding enables output-free llm self-evaluation},
  author={Wang, Yiming and Zhang, Pei and Yang, Baosong and Wong, Derek F and Wang, Rui},
  journal={arXiv preprint arXiv:2410.13640},
  year={2024}
}

@article{du2025latent,
  title={Latent Thinking Optimization: Your Latent Reasoning Language Model Secretly Encodes Reward Signals in Its Latent Thoughts},
  author={Du, Hanwen and Dong, Yuxin and Ning, Xia},
  journal={arXiv preprint arXiv:2509.26314},
  year={2025}
}

@article{chen2024language,
  title={Language models are hidden reasoners: Unlocking latent reasoning capabilities via self-rewarding},
  author={Chen, Haolin and Feng, Yihao and Liu, Zuxin and Yao, Weiran and Prabhakar, Akshara and Heinecke, Shelby and Ho, Ricky and Mui, Phil and Savarese, Silvio and Xiong, Caiming and others},
  journal={arXiv preprint arXiv:2411.04282},
  year={2024}
}

@article{li2025generalist,
  title={Generalist Reward Models: Found Inside Large Language Models},
  author={Li, Yi-Chen and Xu, Tian and Yu, Yang and Zhang, Xuqin and Chen, Xiong-Hui and Ling, Zhongxiang and Chao, Ningjing and Yuan, Lei and Zhou, Zhi-Hua},
  journal={arXiv preprint arXiv:2506.23235},
  year={2025}
}

@article{zhang2023planner,
  title={Planner: Generating diversified paragraph via latent language diffusion model},
  author={Zhang, Yizhe and Gu, Jiatao and Wu, Zhuofeng and Zhai, Shuangfei and Susskind, Joshua and Jaitly, Navdeep},
  journal={Advances in Neural Information Processing Systems},
  volume={36},
  pages={80178--80190},
  year={2023}
}

@article{goyal2023think,
  title={Think before you speak: Training language models with pause tokens},
  author={Goyal, Sachin and Ji, Ziwei and Rawat, Ankit Singh and Menon, Aditya Krishna and Kumar, Sanjiv and Nagarajan, Vaishnavh},
  journal={arXiv preprint arXiv:2310.02226},
  year={2023}
}

@article{hao2024training,
  title={Training large language models to reason in a continuous latent space},
  author={Hao, Shibo and Sukhbaatar, Sainbayar and Su, DiJia and Li, Xian and Hu, Zhiting and Weston, Jason and Tian, Yuandong},
  journal={arXiv preprint arXiv:2412.06769},
  year={2024}
}

@article{geiping2025scaling,
  title={Scaling up test-time compute with latent reasoning: A recurrent depth approach},
  author={Geiping, Jonas and McLeish, Sean and Jain, Neel and Kirchenbauer, John and Singh, Siddharth and Bartoldson, Brian R and Kailkhura, Bhavya and Bhatele, Abhinav and Goldstein, Tom},
  journal={arXiv preprint arXiv:2502.05171},
  year={2025}
}

@article{zhao2023survey,
  title={A survey of large language models},
  author={Zhao, Wayne Xin and Zhou, Kun and Li, Junyi and Tang, Tianyi and Wang, Xiaolei and Hou, Yupeng and Min, Yingqian and Zhang, Beichen and Zhang, Junjie and Dong, Zican and others},
  journal={arXiv preprint arXiv:2303.18223},
  volume={1},
  number={2},
  year={2023}
}

@article{hendrycks2021measuring,
  title={Measuring mathematical problem solving with the math dataset},
  author={Hendrycks, Dan and Burns, Collin and Kadavath, Saurav and Arora, Akul and Basart, Steven and Tang, Eric and Song, Dawn and Steinhardt, Jacob},
  journal={arXiv preprint arXiv:2103.03874},
  year={2021}
}

@article{cobbe2021training,
  title={Training verifiers to solve math word problems},
  author={Cobbe, Karl and Kosaraju, Vineet and Bavarian, Mohammad and Chen, Mark and Jun, Heewoo and Kaiser, Lukasz and Plappert, Matthias and Tworek, Jerry and Hilton, Jacob and Nakano, Reiichiro and others},
  journal={arXiv preprint arXiv:2110.14168},
  year={2021}
}

@article{chen2021evaluating,
  title={Evaluating large language models trained on code},
  author={Chen, Mark},
  journal={arXiv preprint arXiv:2107.03374},
  year={2021}
}

@article{lee2023platypus,
  title={Platypus: Quick, cheap, and powerful refinement of llms},
  author={Lee, Ariel N and Hunter, Cole J and Ruiz, Nataniel},
  journal={arXiv preprint arXiv:2308.07317},
  year={2023}
}

@article{kalamkar2019study,
  title={A study of BFLOAT16 for deep learning training},
  author={Kalamkar, Dhiraj and Mudigere, Dheevatsa and Mellempudi, Naveen and Das, Dipankar and Banerjee, Kunal and Avancha, Sasikanth and Vooturi, Dharma Teja and Jammalamadaka, Nataraj and Huang, Jianyu and Yuen, Hector and others},
  journal={arXiv preprint arXiv:1905.12322},
  year={2019}
}

@article{wen2025reinforcement,
  title={Reinforcement learning with verifiable rewards implicitly incentivizes correct reasoning in base llms},
  author={Wen, Xumeng and Liu, Zihan and Zheng, Shun and Ye, Shengyu and Wu, Zhirong and Wang, Yang and Xu, Zhijian and Liang, Xiao and Li, Junjie and Miao, Ziming and others},
  journal={arXiv preprint arXiv:2506.14245},
  year={2025}
}

@article{zhou2025reinforcing,
  title={Reinforcing General Reasoning without Verifiers},
  author={Zhou, Xiangxin and Liu, Zichen and Sims, Anya and Wang, Haonan and Pang, Tianyu and Li, Chongxuan and Wang, Liang and Lin, Min and Du, Chao},
  journal={arXiv preprint arXiv:2505.21493},
  year={2025}
}

@article{zheng2023judging,
  title={Judging llm-as-a-judge with mt-bench and chatbot arena},
  author={Zheng, Lianmin and Chiang, Wei-Lin and Sheng, Ying and Zhuang, Siyuan and Wu, Zhanghao and Zhuang, Yonghao and Lin, Zi and Li, Zhuohan and Li, Dacheng and Xing, Eric and others},
  journal={Advances in neural information processing systems},
  volume={36},
  pages={46595--46623},
  year={2023}
}

@article{xu2025tinyv,
  title={TinyV: Reducing False Negatives in Verification Improves RL for LLM Reasoning},
  author={Xu, Zhangchen and Li, Yuetai and Jiang, Fengqing and Ramasubramanian, Bhaskar and Niu, Luyao and Lin, Bill Yuchen and Poovendran, Radha},
  journal={arXiv preprint arXiv:2505.14625},
  year={2025}
}

@article{tao2025hybrid,
  title={Hybrid Reinforcement: When Reward Is Sparse, It's Better to Be Dense},
  author={Tao, Leitian and Kulikov, Ilia and Saha, Swarnadeep and Wang, Tianlu and Xu, Jing and Li, Sharon and Weston, Jason E and Yu, Ping},
  journal={arXiv preprint arXiv:2510.07242},
  year={2025}
}

@article{cui2025process,
  title={Process reinforcement through implicit rewards},
  author={Cui, Ganqu and Yuan, Lifan and Wang, Zefan and Wang, Hanbin and Zhang, Yuchen and Chen, Jiacheng and Li, Wendi and He, Bingxiang and Fan, Yuchen and Yu, Tianyu and others},
  journal={arXiv preprint arXiv:2502.01456},
  year={2025}
}

@article{cai2025reinforcement,
  title={Reinforcement learning with verifiable yet noisy rewards under imperfect verifiers},
  author={Cai, Xin-Qiang and Wang, Wei and Liu, Feng and Liu, Tongliang and Niu, Gang and Sugiyama, Masashi},
  journal={arXiv preprint arXiv:2510.00915},
  year={2025}
}

@inproceedings{gao2023scaling,
  title={Scaling laws for reward model overoptimization},
  author={Gao, Leo and Schulman, John and Hilton, Jacob},
  booktitle={International Conference on Machine Learning},
  pages={10835--10866},
  year={2023},
  organization={PMLR}
}

@inproceedings{lightman2023let,
  title={Let's verify step by step},
  author={Lightman, Hunter and Kosaraju, Vineet and Burda, Yuri and Edwards, Harrison and Baker, Bowen and Lee, Teddy and Leike, Jan and Schulman, John and Sutskever, Ilya and Cobbe, Karl},
  booktitle={The Twelfth International Conference on Learning Representations},
  year={2023}
}

@article{zou2023representation,
  title={Representation engineering: A top-down approach to ai transparency},
  author={Zou, Andy and Phan, Long and Chen, Sarah and Campbell, James and Guo, Phillip and Ren, Richard and Pan, Alexander and Yin, Xuwang and Mazeika, Mantas and Dombrowski, Ann-Kathrin and others},
  journal={arXiv preprint arXiv:2310.01405},
  year={2023}
}

@article{bartoszcze2025representation,
  title={Representation Engineering for Large-Language Models: Survey and Research Challenges},
  author={Bartoszcze, Lukasz and Munshi, Sarthak and Sukidi, Bryan and Yen, Jennifer and Yang, Zejia and Williams-King, David and Le, Linh and Asuzu, Kosi and Maple, Carsten},
  journal={arXiv preprint arXiv:2502.17601},
  year={2025}
}

@article{marks2023geometry,
  title={The geometry of truth: Emergent linear structure in large language model representations of true/false datasets},
  author={Marks, Samuel and Tegmark, Max},
  journal={arXiv preprint arXiv:2310.06824},
  year={2023}
}

@article{afzal2025knowing,
  title={Knowing Before Saying: LLM Representations Encode Information About Chain-of-Thought Success Before Completion},
  author={Afzal, Anum and Matthes, Florian and Chechik, Gal and Ziser, Yftah},
  journal={arXiv preprint arXiv:2505.24362},
  year={2025}
}

@article{shai2024transformers,
  title={Transformers represent belief state geometry in their residual stream},
  author={Shai, Adam and Teixeira, Lucas and Oldenziel, Alexander and Marzen, Sarah and Riechers, Paul},
  journal={Advances in Neural Information Processing Systems},
  volume={37},
  pages={75012--75034},
  year={2024}
}

@article{papyan2020prevalence,
  title={Prevalence of neural collapse during the terminal phase of deep learning training},
  author={Papyan, Vardan and Han, XY and Donoho, David L},
  journal={Proceedings of the National Academy of Sciences},
  volume={117},
  number={40},
  pages={24652--24663},
  year={2020},
  publisher={National Academy of Sciences}
}

@article{hendrycks2020measuring,
  title={Measuring massive multitask language understanding},
  author={Hendrycks, Dan and others},
  journal={arXiv preprint arXiv:2009.03300},
  year={2020}
}

@article{hendrycksmath2021,
  title={Measuring Mathematical Problem Solving With the MATH Dataset},
  author={Hendrycks, Dan and others},
  journal={NeurIPS},
  year={2021}
}

@article{suzgun2022challenging,
  title={Challenging BIG-bench tasks and whether chain-of-thought can solve them},
  author={Suzgun, Mirac and others},
  journal={arXiv preprint arXiv:2210.09261},
  year={2022}
}

@article{radford2018improving,
  title={Improving language understanding by generative pre-training},
  author={Radford, Alec and Narasimhan, Karthik and Salimans, Tim and Sutskever, Ilya and others},
  year={2018},
  publisher={San Francisco, CA, USA}
}

\appendix

\section{Usage of LLM}
Large Language Models were used exclusively to improve the clarity and fluency of English writing. They were not involved in research ideation, experimental design, data analysis, or interpretation. The authors take full responsibility for all content.

\section{Detailed Related Work}
\label{app:related}
\noindent\textbf{Policy Optimization and Group-based Variants.}In early RLHF\cite{ziegler2019fine,stiennon2020learning,ouyang2022training, lee2023rlaif} practices, PPO (Proximal Policy Optimization)\cite{schulman2017proximal} was the core algorithm, stabilizing training through trust region constraints. However, PPO requires maintaining a critic model of the same scale as the policy model, introducing significant memory overhead. DPO (Direct Preference Optimization)\cite{rafailov2023direct} subsequently eliminated the need for critics by reparameterizing rewards as functions of optimal policies, but its offline nature limits exploration capabilities. To balance online exploration and computational efficiency, GRPO (Group Relative Policy Optimization)\cite{shao2024deepseekmath} introduced group-based advantage estimation and has become a research focus. A series of variants have emerged to address specific challenges: DAPO\cite{yu2025dapo} combines dynamic sampling with gradient clipping for training stability, GSPO\cite{zheng2025group} employs sequence-level importance sampling for MoE\cite{jacobs1991adaptive,du2022glam} models, Dr. GRPO\cite{liu2025understanding} corrects length and difficulty biases, and GMPO\cite{zhao2025geometric} uses geometric mean to resist reward outliers. Despite significant progress at the optimizer level, these methods have not addressed the source of reward signals, still assuming the existence of perfect oracles (ground truth or expensive external verifiers). Latent-GRPO is orthogonal to these works: by extracting intrinsic geometric signals from the model, our reward mechanism can be seamlessly integrated with any of these optimization algorithms, achieving both ``no external supervision'' and ``efficient stable optimization.''

\noindent\textbf{Training-free evaluation Methods.}To eliminate dependence on expensive labeled data, researchers have explored training-free evaluation and self-correction methods. Self-Consistency (SC)\cite{wang2022self}  improves answer accuracy through majority voting over multiple reasoning paths. Self-Refine\cite{madaan2023self} enables iterative refinement by generating self-feedback without external guidance. Tree-of-Thoughts (ToT)\cite{yao2023tree} models problem-solving as tree search, exploring multiple reasoning branches and backtracking to find optimal solutions. Best-of-N\cite{stiennon2020learning} sampling generates multiple candidates and selects the best based on heuristics or confidence scores. Forest-of-Thoughts\cite{bi2024forest} extends ToT by integrating multiple trees with sparse activation for efficiency. Speculative Rejection\cite{sun2024fast} accelerates Best-of-N by early termination of low-quality candidates. While these methods leverage various forms of ``consensus'' or ``selection,'' they primarily rely on discrete matching of final text results or explicit model outputs, failing to capture subtle semantic differences in reasoning processes. Moreover, they are typically used only during inference without converting to training signals. Unlike these approaches, Latent-GRPO does not rely on explicit text feedback but uses geometric structures in the latent space as implicit evaluation criteria. We demonstrate that geometric centroids in latent space contain richer semantic information than explicit text-based methods, providing denser and more robust gradient signals for RL training.

\noindent\textbf{Latent Space and Latent Thinking.}The latent space of LLMs is widely recognized to encode rich semantic and structural information\cite{zhang2023planner,goyal2023think,hao2024training,geiping2025scaling}. Early probing work revealed that LLM hidden states contain linear directions representing syntax, sentiment, and truthfulness. Recent work has explored latent thinking, replacing verbose natural language steps with compact latent representations. CoE (Chain-of-Embedding)\cite{wang2024latent} uses progressive hidden states for output-free self-evaluation, showing significant differences between correct and incorrect answers without requiring training. LTO (Latent Thinking Optimization)\cite{du2025latent} and LaTRO\cite{chen2024language} train additional latent classifiers (Latent Reward Models) to predict correctness and guide RL optimization, achieving 12.5\% average accuracy improvement on GSM8K. However, these methods require training extra models, introducing computational overhead. EndoRM\cite{li2025generalist} extracts ``endogenous rewards'' from LLM logits without training, but relies on output probabilities rather than latent geometric structures. Few works have attempted to directly convert geometric consensus in latent space into RL reward signals without additional training. Our work fills this gap by demonstrating that high-quality reward signals can be extracted from latent space through simple geometric methods. We establish that ``Last Token Hidden States'' combined with ``Iterative Robust Centroid Estimation'' effectively captures subtle semantic differences in reasoning, providing denser and more robust gradient signals than explicit text-based or probability-based methods.

\section{Detailed Analysis of Geometric-Based Scoring}
\label{app:section3ana}
We simulate the actual GRPO training scenario to validate whether geometric features can serve as reliable quality indicators. For each prompt, we generate a group of $G=8$ trajectories (matching the typical GRPO group size) and compute geometric scores using our Iterative Robust Centroid Estimation (IRCE) algorithm. Figure~\ref{fig:correlation} presents the effectiveness of geometric-based scoring:

\textbf{Distribution Separability (Figure~\ref{fig:correlation}a).} The box plot analysis reveals clear separation between different quality levels in the geometric reward space. Trajectories with high external scores (green, score $> 0.9$) exhibit highly concentrated distributions with median geometric scores close to 0 (near the centroid), while low-score trajectories (red, score $< 0.3$) show significantly right-skewed distributions with large variance (far from the centroid). The partial-correct trajectories (yellow, $0.3 < $ score $< 0.9$) fall in between, demonstrating that geometric features provide fine-grained quality discrimination rather than binary classification.

\textbf{Group-Level Ranking Consistency (Figure~\ref{fig:correlation}b-d).} To validate the practical utility in GRPO's advantage estimation, we examine three representative groups of 8 trajectories. The dual-axis plot shows ground truth scores (gray bars) and normalized geometric scores (blue line with markers). The Spearman rank correlation of 0.927 indicates high ranking consistency between geometric and external scores in the Figure~\ref{fig:correlation}b. Critically, the top-ranked samples identified by geometric scores (red circles) align well with ground truth, demonstrating that our method can effectively identify elite trajectories for policy optimization. In the Top-1 selection task across all groups, the consistency rate between geometric-based selection and external model selection exceeds 85\%, confirming that geometric signals can effectively replace expensive external verifiers in GRPO's advantage estimation.

\section{Detailed Introduction of Iterative Robust Centroid Estimation}
\label{app:irce}

For each prompt, the policy generates a group of $G$ trajectories. From each trajectory, we extract the \textbf{terminal hidden state} $\mathbf{h}_i \in \mathbb{R}^d$ ($i=1, \dots, G$), forming the sampled state matrix $\mathbf{H} = [\mathbf{h}_1, \dots, \mathbf{h}_G]^\top \in \mathbb{R}^{G \times d}$. The Iterative Robust Centroid Estimation (IRCE) algorithm proceeds through the following steps:

\textbf{Step 1: Spherical Projection.} 
To eliminate magnitude fluctuations and ensure the distance metric focuses exclusively on semantic directionality, we apply $L_2$ normalization to project all terminal hidden states onto a unit hypersphere:
\begin{equation}
\tilde{\mathbf{h}}_i = \frac{\mathbf{h}_i}{\|\mathbf{h}_i\|_2}
\end{equation}

\textbf{Step 2: Initialize Centroid.} 
The initial centroid $\boldsymbol{\mu}^{(0)}$ is computed as the normalized mean of the projected states:
\begin{equation}
\boldsymbol{\mu}^{(0)} = \frac{\frac{1}{G} \sum_{i=1}^{G} \tilde{\mathbf{h}}_i}{\left\| \frac{1}{G} \sum_{i=1}^{G} \tilde{\mathbf{h}}_i \right\|_2}
\end{equation}

\textbf{Step 3: Iterative Soft-Weighted Update.} 
In each iteration $s = 0, 1, \dots, T-1$, we compute the Euclidean distance from each sample to the current centroid:
\begin{equation}
d_i^{(s)} = \|\tilde{\mathbf{h}}_i - \boldsymbol{\mu}^{(s)}\|_2
\end{equation}
To adapt to varying cluster densities across different training stages, we derive an adaptive scale parameter $\sigma^{(s)}$ based on the group distance distribution:
\begin{equation}
\sigma^{(s)} = \text{std}(\{d_1^{(s)}, \dots, d_G^{(s)}\}) + \epsilon
\end{equation}
where $\epsilon$ is a small constant for numerical stability. We then calculate soft weights $w_i^{(s)}$ via a Gaussian kernel to diminish the influence of outliers:
\begin{equation}
w_i^{(s)} = \frac{\exp\left(-\frac{(d_i^{(s)})^2}{2(\sigma^{(s)})^2}\right)}{\sum_{j=1}^G \exp\left(-\frac{(d_j^{(s)})^2}{2(\sigma^{(s)})^2}\right)}
\end{equation}
The centroid is updated using a weighted average and subsequently re-projected onto the hypersphere:
\begin{equation}
\boldsymbol{\mu}^{(s+1)} = \frac{\sum_{i=1}^G w_i^{(s)} \tilde{\mathbf{h}}_i}{\left\| \sum_{i=1}^G w_i^{(s)} \tilde{\mathbf{h}}_i \right\|_2}
\end{equation}

\textbf{Step 4: Reward Computation.} 
Upon convergence after $T$ iterations, the intrinsic reward $R_i$ for each trajectory is defined as the negative Euclidean distance to the final centroid:
\begin{equation}
R_i = -\|\tilde{\mathbf{h}}_i - \boldsymbol{\mu}^{(T)}\|_2
\end{equation}
Optionally, we apply Min-Max normalization within the group to scale the rewards $R \in [0, 1]$, ensuring stable gradient estimation for the policy update.

\section{Usage of Datasets}
\label{app:datasets}
\subsection{Datasets Overview}
We use three main datasets for our experiments, each serving different purposes in evaluating the generalization and robustness of Latent-GRPO. Table~\ref{tab:datasets_overview} provides a comprehensive overview of all three datasets, including their sources, sizes, and usage in our RL training.

\begin{table}[t]
\small
\centering
\begin{tabular*}{\columnwidth}{@{\extracolsep{\fill}}lrrr@{}}
\toprule
Dataset & Total Size & Train & Val \\
\midrule
GSM8K & 8,500 & 2,000 & 1,000 \\
MATH & 12,500 & 3,000 & 1,000 \\
Open-Platypus & 20,726 & 4,500 & 2,000 \\
\midrule
\multicolumn{4}{l}{\textit{Open-Platypus Data Source Statistics:}} \\
MATH/PRM-800K & 12,298 & 754 & 783 \\
ReClor & 4,530 & 693 & 437 \\
Airoboros & 2,605 & 528 & 251 \\
ScienceQA & 1,317 & 618 & 127 \\
LeetCode Solutions & 1,100 & 365 & 106 \\
ARB & 713 & 272 & 69 \\
SciBench & 616 & 249 & 59 \\
TheoremQA & 564 & 336 & 54 \\
TigerBot Kaggle & 386 & 193 & 37 \\
OpenAssistant Guanaco & 797 & 492 & 77 \\
Total & 20,726 & 4,500 & 2,000 \\
\bottomrule
\end{tabular*}
\caption{Overview of datasets used in RL training}
\label{tab:datasets_overview}
\end{table}

\subsection{Dataset Details}

GSM8K is a dataset of 8,500 grade school math word problems designed to test multi-step mathematical reasoning. MATH is a challenging dataset containing 12,500 competition-level mathematics problems from high school and undergraduate mathematics competitions. Open-Platypus is a large-scale diverse instruction-following dataset focused on improving LLM logical reasoning skills, comprising 20,726 samples filtered from multiple sources (MATH/PRM-800K: 12,298, ReClor: 4,530, Airoboros: 2,605, ScienceQA: 1,317, LeetCode Solutions: 1,100, ARB: 713, SciBench: 616, TheoremQA: 564, TigerBot Kaggle: 386, OpenAssistant Guanaco: 797) using keyword search and Sentence Transformers (removing questions with similarity above 80\%).

\subsection{Benchmark}
\label{app:benchmark}
To comprehensively evaluate the performance of our model across general knowledge, logical reasoning, and advanced mathematical problem-solving, we select five representative benchmarks. These range from broad multi-task knowledge to extremely challenging competitive mathematics.

\paragraph{MMLU} Massive Multitask Language Understanding \cite{hendrycks2020measuring} covers 57 subjects across STEM, the humanities, social sciences, and more. It tests both world knowledge and problem-solving capsules.

\paragraph{MATH-500} The MATH dataset \cite{hendrycksmath2021} consists of 12,500 challenging high school math competition problems. Following recent literature (e.g., DeepSeek-R1), we use the \textbf{MATH-500} subset, which comprises 500 representative problems from the test set to evaluate rigorous mathematical reasoning.

\paragraph{BBH} Big-Bench Hard (BBH) \cite{suzgun2022challenging} focuses on a subset of 23 challenging tasks from the BIG-bench suite where previous language models fell short of human multi-step reasoning capabilities.

\paragraph{AIME 24 \& AIME 25} The American Invitational Mathematics Examination (AIME) is a prestigious high school competition. We evaluate our model on the \textbf{AIME 2024} and the most recent \textbf{AIME 2025} problems. These sets are particularly valuable as "out-of-distribution" (OOD) tests for reasoning models, as they require creative multi-step logical chains without standard templates.

\begin{table*}[t] 
\centering
\small 
\setlength{\tabcolsep}{0pt} 
\begin{tabular*}{\textwidth}{@{\extracolsep{\fill}}llrl@{}}
\toprule
\textbf{Benchmark} & \textbf{Domain} & \textbf{Size} & \textbf{Metric} \\ 
\midrule
MMLU & General Knowledge (57 subjects) & 14,042 & 5-shot Accuracy \\
MATH-500 & Competition Level Mathematics & 500 & Pass@1 \\
BBH & Logical \& Multi-step Reasoning & 6,511 & 3-shot / CoT Accuracy \\
AIME 24 & Advanced Mathematics Competition & 30 & Pass@k / Consensual \\
AIME 25 & Advanced Mathematics Competition & 30 & Pass@k / Consensual \\ 
\bottomrule
\end{tabular*}
\caption{\label{tab:benchmarks} Summary of evaluation benchmarks used in this work. For reasoning tasks, we prioritize pass@k and consensual evaluation to measure thinking capabilities.}
\end{table*}

\section{Experimental Hyperparameters}
\label{app:hyperparameters}

\subsection{Hardware Configuration}
All experiments are conducted on 1  GPU using PyTorch 2.6, CUDA 12.4 and the Transformers library 4.51.1. We employ bfloat16\cite{kalamkar2019study} mixed precision training to improve memory efficiency and computational speed. 

\subsection{GRPO Training Configuration}

All models use the Instruct version with Flash Attention 2 enabled. Maximum sequence length is 8192, generation temperature is 0.95, and Top-p sampling is 0.9. Group size is 8 trajectories per prompt, batch size is 1 prompt with mini-batch size 2. PPO inner update rounds are 1, clip ratio is 0.2, KL penalty coefficient is 0.1. Learning rate is $1 \times 10^{-5}$ with linear warmup over 100 steps, maximum gradient norm is 1.0, weight decay is 0.01. We train for 2 epochs with logging every 10 steps and checkpoints every 100 steps.  To simulate scenarios with constrained computational resources, the throughput for LLM-as-Judge scoring is restricted to 2 Queries Per Second (QPS).

\subsection{IRCE Algorithm Configuration}

The Iterative Robust Centroid Estimation algorithm uses 5 iterations with temperature parameter 0.5. Rewards are Min-Max normalized to $[0, 1]$ interval. Convergence threshold is $1 \times 10^{-6}$ for early stopping. Epsilon for numerical stability is $1 \times 10^{-8}$.

\section{Detailed Reward Methods}
\label{app:method}

\subsection{Basemethods of reward}
\label{app:basemethod}
\noindent\textbf{Rule-based reward} For mathematical problems, we use symbolic computation (SymPy) to verify correctness by comparing the model's final answer with the ground truth. For code generation tasks, we use sandboxed execution environments to test whether the generated code produces correct outputs. Rule-based reward provides binary rewards (0 or 1) based on exact correctness, making it a reliable but sparse reward signal. The main limitation is that it requires well-defined verification rules and cannot be applied to open-ended tasks like writing or summarization.

\noindent\textbf{LLM-as-Judge.} We use GPT-4o to evaluate response quality with a standardized prompt asking for binary correctness judgment. Each response is evaluated independently, and the judgment is converted to a binary reward (0 or 1). While LLM-as-Judge is more flexible and can handle diverse task types, it introduces external dependency, higher computational cost, and potential inconsistency in evaluation. 

\subsection{Latent Reward Methods}
\label{app:latent_reward}
We compare four latent reward methods that extract rewards from hidden states without external supervision:

\noindent\textbf{Mean Pool.} Computes the centroid as the simple arithmetic mean of all normalized hidden states: $\boldsymbol{\mu} = \frac{1}{G} \sum_{i=1}^{G} \tilde{\mathbf{h}}_i$. Rewards are computed as negative distances: $R_i = -||\tilde{\mathbf{h}}_i - \boldsymbol{\mu}||_2$. This method is computationally efficient but sensitive to outliers. 

\noindent\textbf{K-Means.} Applies K-Means clustering with $K=2$ to group hidden states into quality and non-quality clusters. The algorithm iteratively optimizes cluster centers and assigns samples to nearest centers. Rewards are based on distance to the quality cluster center. This method identifies dominant quality patterns but requires parameter tuning and has higher computational overhead. 

\noindent\textbf{Eigen Centrality.} Constructs a similarity matrix $\mathbf{A}$ based on cosine similarity between hidden states. The principal eigenvector $\mathbf{v}$ of $\mathbf{A}$ is computed, and component $v_i$ serves as the reward for sample $i$. This method captures global graph structure but suffers from eigendecomposition computational bottleneck. 

\noindent\textbf{IRCE (Ours).} Performs iterative soft-weighted centroid estimation. Step 1 normalizes hidden states to unit hypersphere. Step 2 initializes centroid as mean of normalized states. Step 3 iteratively updates centroid using soft weights based on Gaussian kernel. Step 4 computes final rewards as negative distances. 

\section{Detailed Analysis of Experiments}
\label{app:analysis}
\begin{figure*}[t]
\centering

\begin{subfigure}{0.48\linewidth}
  \centering
  \fbox{\includegraphics[width=\linewidth]{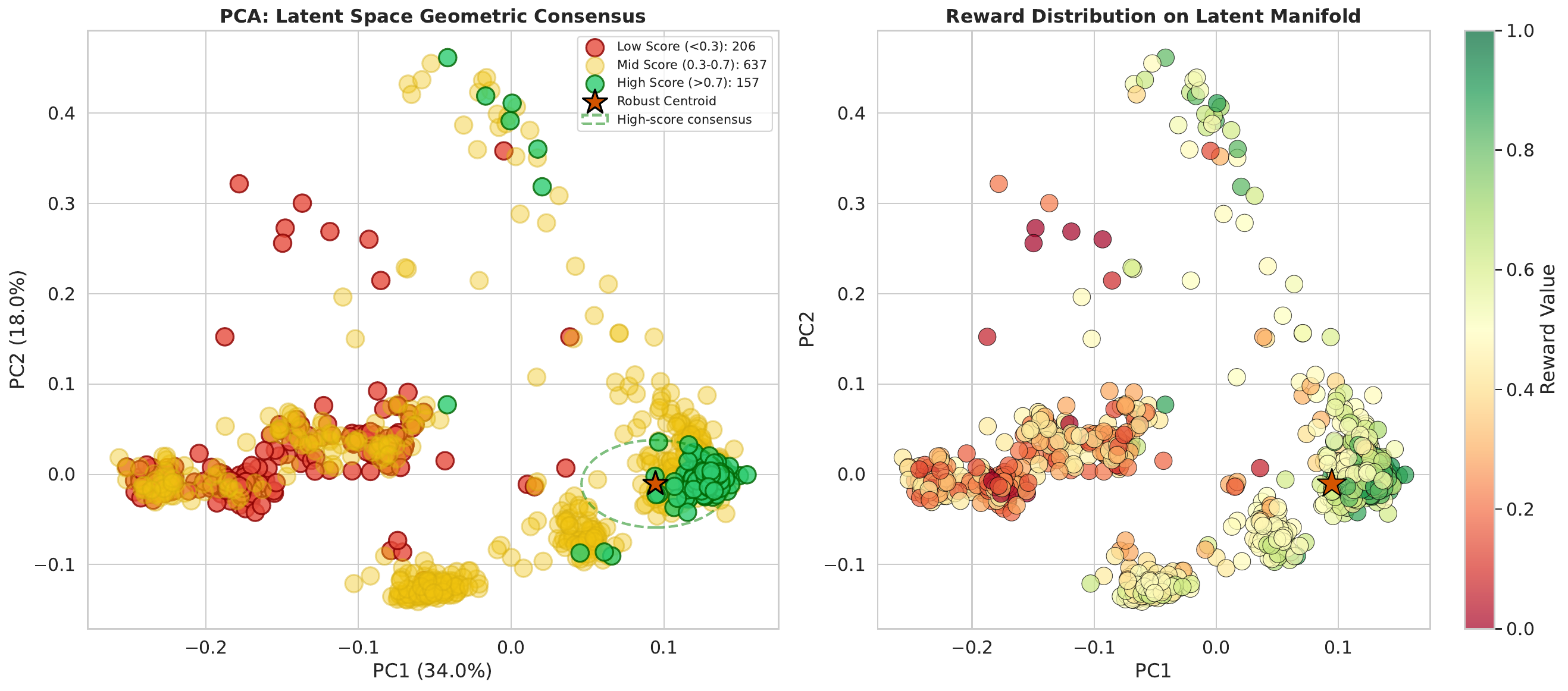}}
  \caption{Clustering of Qwen3-1.7B's 1000 rollouts on ScienceQA}
  \label{fig:1.7_science}
\end{subfigure}
\hfill
\begin{subfigure}{0.48\linewidth}
  \centering
  \fbox{\includegraphics[width=\linewidth]{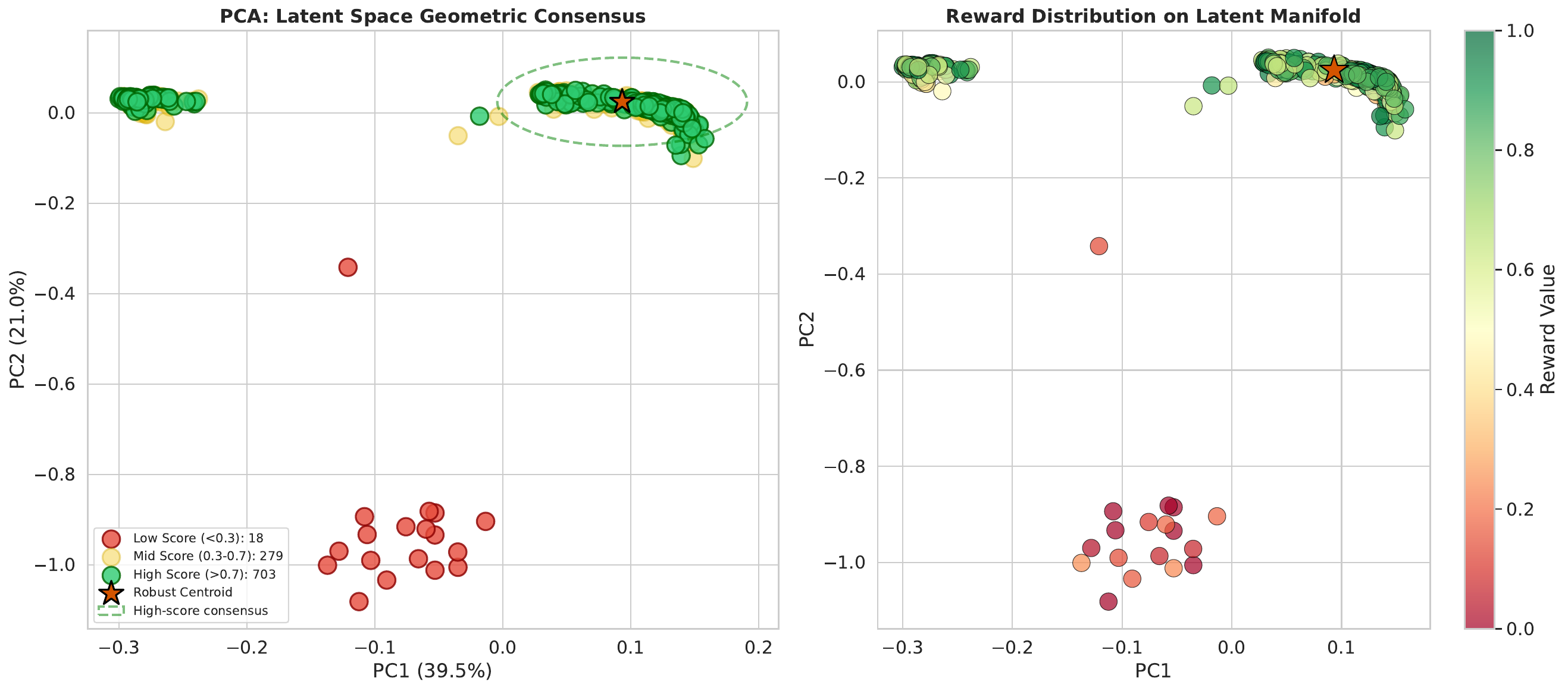}}
  \caption{Clustering of Qwen3-4B's 1000 rollouts on ARBdata}
  \label{fig:4_arb}
\end{subfigure}

\vspace{1.5em} 

\begin{subfigure}{0.48\linewidth}
  \centering
  \fbox{\includegraphics[width=\linewidth]{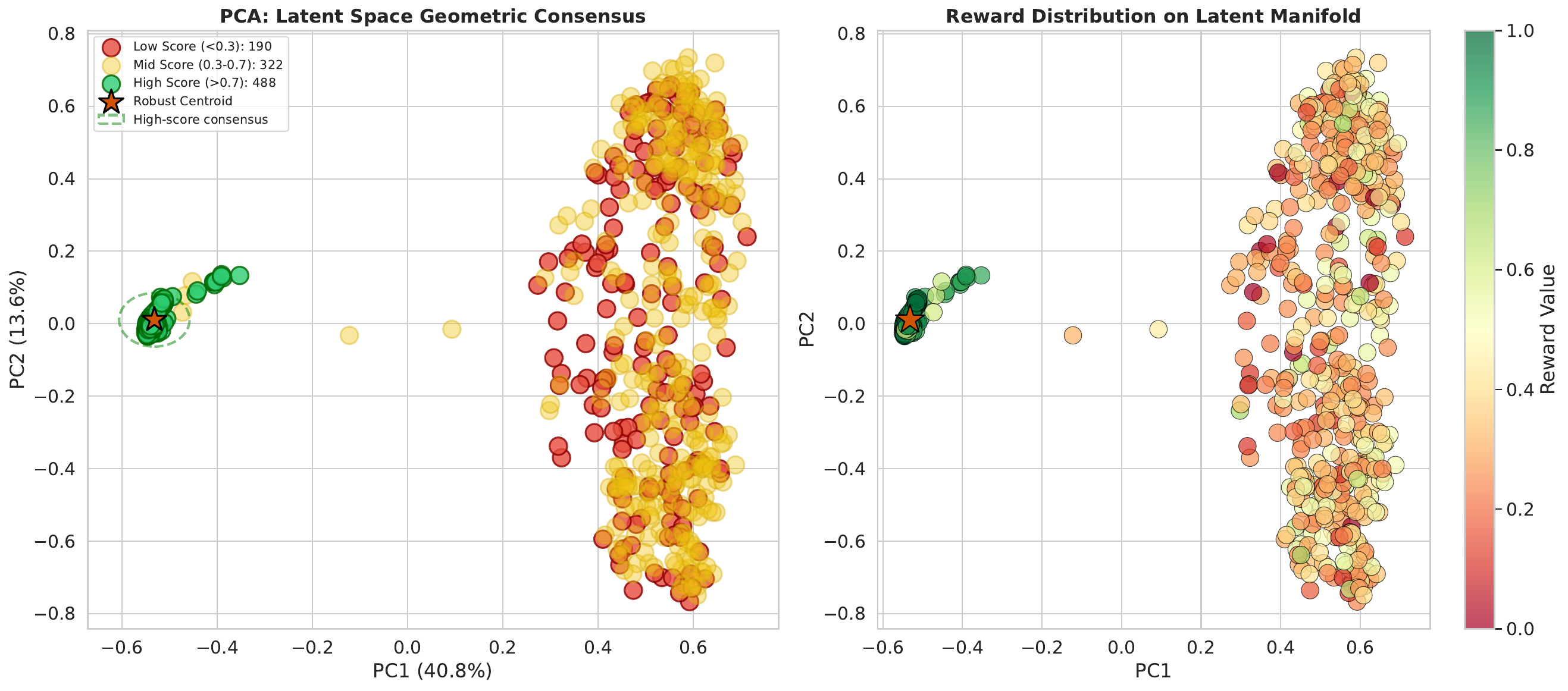}}
  \caption{Clustering of the Qwen3-4B's 1000 rollouts on Mathdata}
  \label{fig:4_math}
\end{subfigure}
\hfill
\begin{subfigure}{0.48\linewidth}
  \centering
  \fbox{\includegraphics[width=\linewidth]{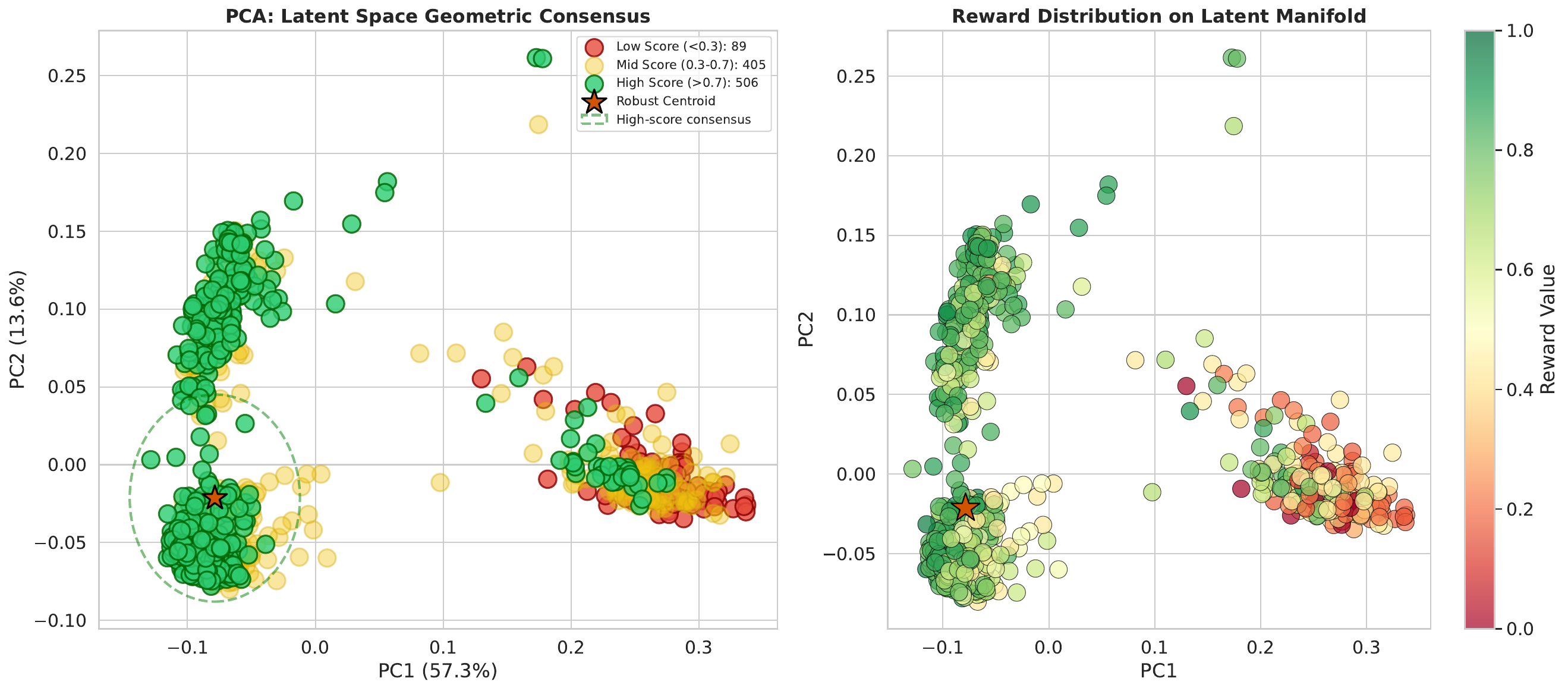}}
  \caption{Clustering of the Qwen3-4B's 1000 rollouts on ScienceQA}
  \label{fig:4_science}
\end{subfigure}

\caption{Visualization of latent manifold consensus across various model scales and benchmarks. Each panel shows the 2D PCA projection of 1,000 terminal hidden states. (a-d) consistently demonstrate that correct trajectories (green) cluster into a dense consensus core, while incorrect ones (red) are scattered.}
\label{fig:all_clustering_results}
\end{figure*}

\subsection{Main Experiments}
\label{app:main}
\noindent\textbf{On GSM8K:} Latent-GRPO achieves competitive or superior accuracy compared to LLM-as-Judge across all model scales. On Qwen-0.6B, Latent-GRPO reaches 61.25\% versus 53.52\% for LLM-as-Judge, while significantly reducing training time from 768.42\,m to 431.18\,m per epoch (\textbf{1.78$\times$ speed-up}). On Qwen-1.7B, Latent-GRPO outperforms LLM-as-Judge by a substantial margin (73.88\% vs 64.20\%) with \textbf{2.10$\times$ speed-up} (1032.55\,m vs 492.34\,m). On Qwen-4B, Latent-GRPO achieves 82.34\% accuracy with \textbf{2.14$\times$ speed-up} (1411.72\,m vs 658.21\,m). Compared to Rule-based verification, Latent-GRPO consistently achieves higher accuracy while maintaining competitive training efficiency.

\noindent\textbf{On MATH:} Latent-GRPO consistently outperforms LLM-as-Judge across all model scales with substantial speed-ups. On Qwen-0.6B, Latent-GRPO improves from 52.94\% to 58.47\% while reducing time from 1224.15\,m to 718.63\,m (\textbf{1.70$\times$ speed-up}). On Qwen-1.7B, Latent-GRPO achieves 78.51\% versus 65.77\% with \textbf{1.98$\times$ speed-up} (1608.34\,m vs 811.51\,m). On Qwen-4B, Latent-GRPO achieves 77.53\% versus 77.44\% with \textbf{2.18$\times$ speed-up} (2357.31\,m vs 1081.47\,m), demonstrating that even when accuracy is comparable, significant computational-efficiency gains are achieved. Rule-based methods show weakness on Qwen-1.7B (42.14\%), where Latent-GRPO substantially outperforms.

\noindent\textbf{On Open-Platypus:} Latent-GRPO demonstrates the strongest improvements on this diverse-reasoning benchmark. On Qwen-0.6B, Latent-GRPO achieves 40.56\% versus 34.45\% with \textbf{1.80$\times$ speed-up} (1937.82\,m vs 1079.27\,m). On Qwen-1.7B, Latent-GRPO reaches 64.82\% versus 56.69\% with \textbf{2.11$\times$ speed-up}. Most notably, on Qwen-4B, Latent-GRPO achieves 78.06\% versus 65.21\% with the highest speed-up of \textbf{2.16$\times$} (3522.18\,m vs 1632.52\,m), demonstrating that intrinsic rewards are particularly effective for complex, diverse reasoning tasks where external verification becomes a critical computational bottleneck.

\subsection{Ablation Experiments}
\label{app:abl}
The ablation experiments systematically evaluate each component of Latent-GRPO through two critical design choices: hidden state extraction methods and latent score estimation algorithms.

\noindent\textbf{Hidden State Extraction Methods} As shown in Table~\ref{tab:extraction}, Last Token consistently achieves the highest accuracy across all three model scales: 61.25\% on Qwen3-0.6B (2.51\% improvement over Mean Pooling), 73.88\% on Qwen3-1.7B (2.83\% improvement), and 82.34\% on Qwen3-4B (2.89\% improvement). Notably, Weighted Mean (which selectively pools pre-designed key tokens) fails to substantially improve over Mean Pooling on any model scale (57.12\% vs 58.74\% on Qwen3-0.6B, 69.88\% vs 71.05\% on Qwen3-1.7B, 78.12\% vs 79.45\% on Qwen3-4B), suggesting that curating specific keyword positions cannot effectively identify which tokens carry reasoning quality signals.

\noindent\textbf{Latent Score Methods} As shown in Table~\ref{tab:latent_score}, IRCE achieves the highest accuracy on all three model scales: 61.25\% on Qwen3-0.6B (outperforming Mean Pool by 4.13\%), 73.88\% on Qwen3-1.7B (outperforming Mean Pool by 5.43\%), and 82.34\% on Qwen3-4B (outperforming Mean Pool by 4.45\%). IRCE also achieves the best computational efficiency, with inference times of 431.18m (0.6B), 492.34m (1.7B), and 658.21m (4B), substantially faster than K-Means (489.12m, 543.89m, 725.67m) and Eigen Centrality (468.76m, 531.42m, 708.45m).

\section{Generalization of Latent-GRPO}
\label{app:generalization}
\begin{figure*}[ht]
\centering
\includegraphics[width=1\textwidth]{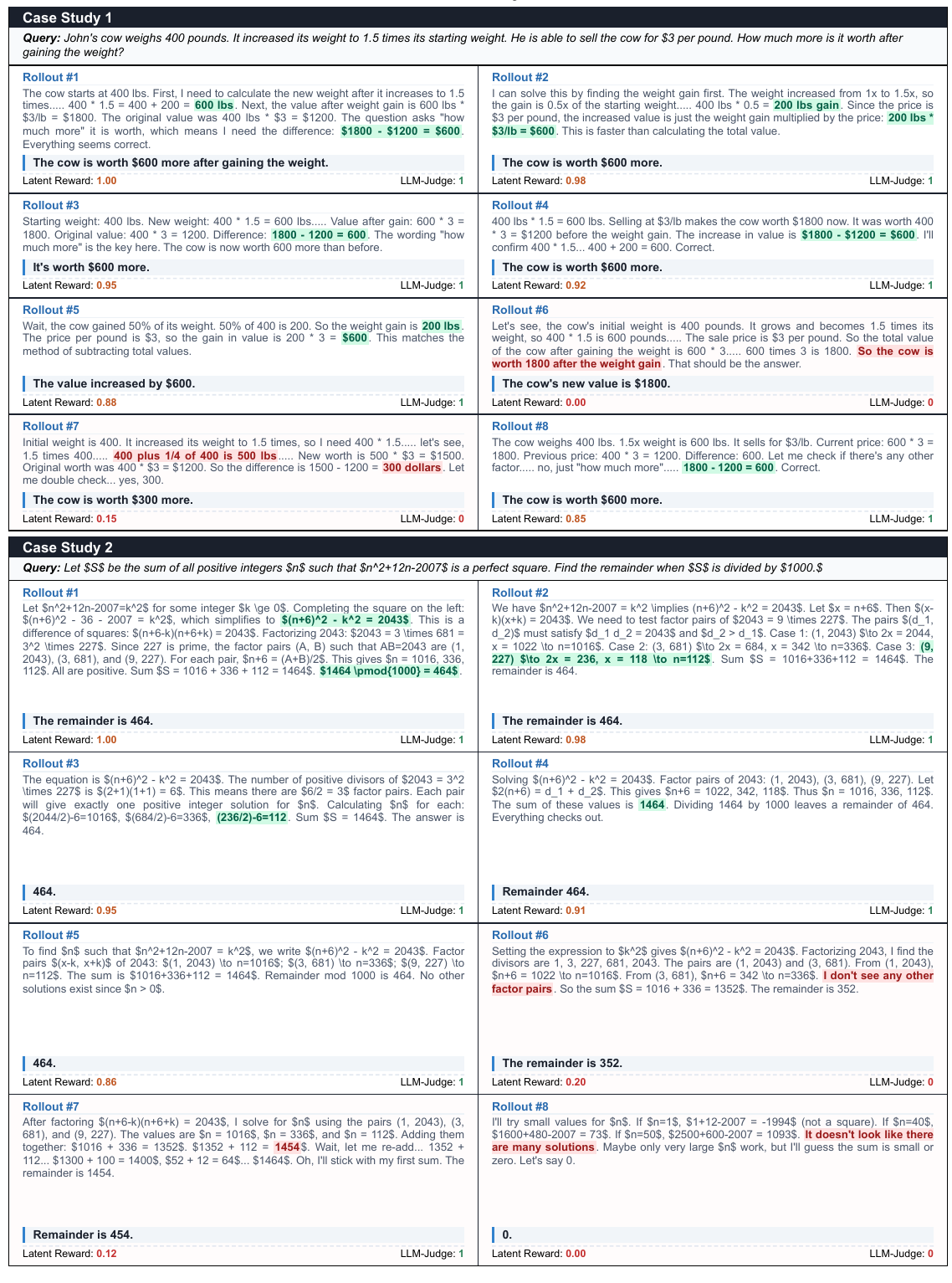}
\caption{Case study of the reward between Latent-GRPO and LLM-as-Judge}
\label{fig:example}
\end{figure*}

\subsection{Generalization of Geometric Properties}
\label{app:Geometric}
To evaluate the generalization of Geometric Properties, we choose additional data, including scientific, logical reasoning, and more challenging mathematical problems. Meanwhile we use models of more sizes to evaluate them. The geometric clustering in latent space demonstrates consistent patterns across different model sizes and datasets.
Each visualization is based on 1,000 rollouts from a single test sample, with correctness scores assigned by GPT-4o. Correct trajectories (green circles, score $>0.7$) form a visibly tight consensus core around the truth centroid (gold star), while incorrect trajectories (red circles, score $<0.3$) scatter as outliers.
The dashed ellipse in each panel outlines the 95\% confidence region of the consensus core. In the original high-dimensional space, incorrect samples consistently lie farther from the centroid than correct ones, quantifying the geometric separability.This qualitative property generalizes across different model scales and benchmarks.
As shown in Figure~\ref{fig:all_clustering_results}, whether on Qwen3-1.7B with ScienceQA or on Qwen3-4B with ARB and Math data, the correct cluster remains compact while the incorrect cloud disperses, indicating that geometric proximity in latent space is a robust proxy for reasoning quality.
The separation pattern is stable across configurations, suggesting that the underlying geometric structure is universal rather than task- or model-specific.

\section{Case study}
Figure ~\ref{fig:example} provides the two cases between Latent-GRPO and LLM-as-Judge method. These two representative cases demonstrate that our Latent-GRPO approach provides accurate continuous reward signals during training. These signals are endogenous in nature, requiring no external verifier supervision, and crucially, they are continuous rather than binary like LLM-as-Judge. This continuous signal not only accelerates iteration efficiency through enhanced logical consistency, but also stabilizes RL training by eliminating model collapse caused by external verifier inconsistency and mitigating training loss from over-reliance on external verifier accuracy.

\end{document}